\documentclass[10pt,twocolumn,letterpaper]{article}

\usepackage{cvpr}
\usepackage{times}
\usepackage{epsfig}
\usepackage{graphicx}
\usepackage{amsmath}
\usepackage{amssymb}
\usepackage{url}
\usepackage{algorithm}
\usepackage{algpseudocode}
\usepackage{float}
\usepackage{booktabs}
\usepackage{multirow}
\usepackage{anyfontsize}
\usepackage{subcaption}
\usepackage{wrapfig}
\usepackage{placeins}
\usepackage{bbm}
\usepackage{amsmath,amsfonts,bm}

\def\eqref#1{equation~\ref{#1}}
\def\1{\bm{1}}

\def\vg{{\bm{g}}}

\def\vr{{\bm{r}}}

\def\vw{{\bm{w}}}
\def\vx{{\bm{x}}}
\def\vy{{\bm{y}}}
\def\vz{{\bm{z}}}

\DeclareMathAlphabet{\mathsfit}{\encodingdefault}{\sfdefault}{m}{sl}
\SetMathAlphabet{\mathsfit}{bold}{\encodingdefault}{\sfdefault}{bx}{n}

\def\gL{{\mathcal{L}}}

\def\gS{{\mathcal{S}}}
\def\gT{{\mathcal{T}}}

\def\sX{{\mathbb{X}}}
\def\sY{{\mathbb{Y}}}

\newcommand{\E}{\mathbb{E}}

\newcommand{\softmax}{\mathrm{softmax}}

\DeclareMathOperator*{\argmax}{arg\,max}
\DeclareMathOperator*{\argmin}{arg\,min}

\usepackage[pagebackref=true,breaklinks=true,letterpaper=true,colorlinks,bookmarks=false]{hyperref}

\cvprfinalcopy % *** Uncomment this line for the final submission

\def\cvprPaperID{8551} % *** Enter the CVPR Paper ID here

\ifcvprfinal\fi
\begin{document}

%%%%%%%%% TITLE
\title{\vspace{-4mm}Neural Data Server: A Large-Scale Search Engine for Transfer Learning Data}

\author{
	Xi Yan$^{1, 2}$\thanks{authors contributed equally}
	\hspace{2cm}
	David Acuna$^{1,2,3}$\footnotemark[1]
	\hspace{2cm}
	Sanja Fidler$^{1,2,3}$\\
$^1$University of Toronto \hspace{1em} $^2$Vector Institute \hspace{1em}  \hspace{1em} $^3$NVIDIA\\
{\tt\small {\tt\small xi.yan@mail.utoronto.ca}, \{davidj, fidler\}@cs.toronto.edu}
}

\newcommand{\DA}[1]{{\color{red}{[David: #1]}}}
\newcommand{\XY}[1]{{\color{red}{[Xi: #1]}}}
\newcommand{\SF}[1]{{\color{magenta}{[SF: #1]}}}

\newcommand{\fix}{\marginpar{FIX}}
\newcommand{\new}{\marginpar{NEW}}
\newcommand{\ds}{\emph{dataserver}}
\newcommand{\cl}{\emph{client}}

\captionsetup{font=footnotesize}

\maketitle
\thispagestyle{empty}

%%%%%%%%% ABSTRACT
\begin{abstract}
Transfer learning has proven to be a successful technique to train 
deep learning models in the domains where little training data is available. 
The dominant approach is to pretrain a model on a large generic dataset such as ImageNet and finetune its weights on the target domain. 
However, in the new era of an ever increasing number of massive datasets, \emph{selecting the relevant data for pretraining} is a critical issue. 
We introduce Neural Data Server (NDS), a large-scale search engine for finding the most useful transfer learning data to the target domain. 
NDS consists of a {\ds} which indexes several large popular image datasets, and aims to recommend data to a {\cl}, an end-user with a target application with its own small labeled dataset. 
The {\ds} represents large datasets with a much more compact mixture-of-experts model, and employs it to perform data search in a series of {\ds}-{\cl} transactions at a low computational cost. We show the effectiveness of NDS in various transfer learning scenarios, demonstrating state-of-the-art performance on several target datasets and tasks such as image classification, object detection and instance segmentation. Neural Data Server is available as a web-service at \url{http://aidemos.cs.toronto.edu/nds/}. 

\end{abstract}

%%%%%%%%% BODY TEXT
\vspace{-5mm}
\section{Introduction}
\vspace{-1.5mm}
In recent years, we have seen an explosive growth of the number and the variety of computer vision applications. These range from generic image classification tasks to surveillance, sports analytics, clothing recommendation, early disease detection, and to mapping, among others. Yet, we are only at the beginning of our exploration of what is possible to achieve with Deep Learning.

\begin{figure}[t]
\vspace{-1mm}
\begin{center}
\includegraphics[width=\linewidth]{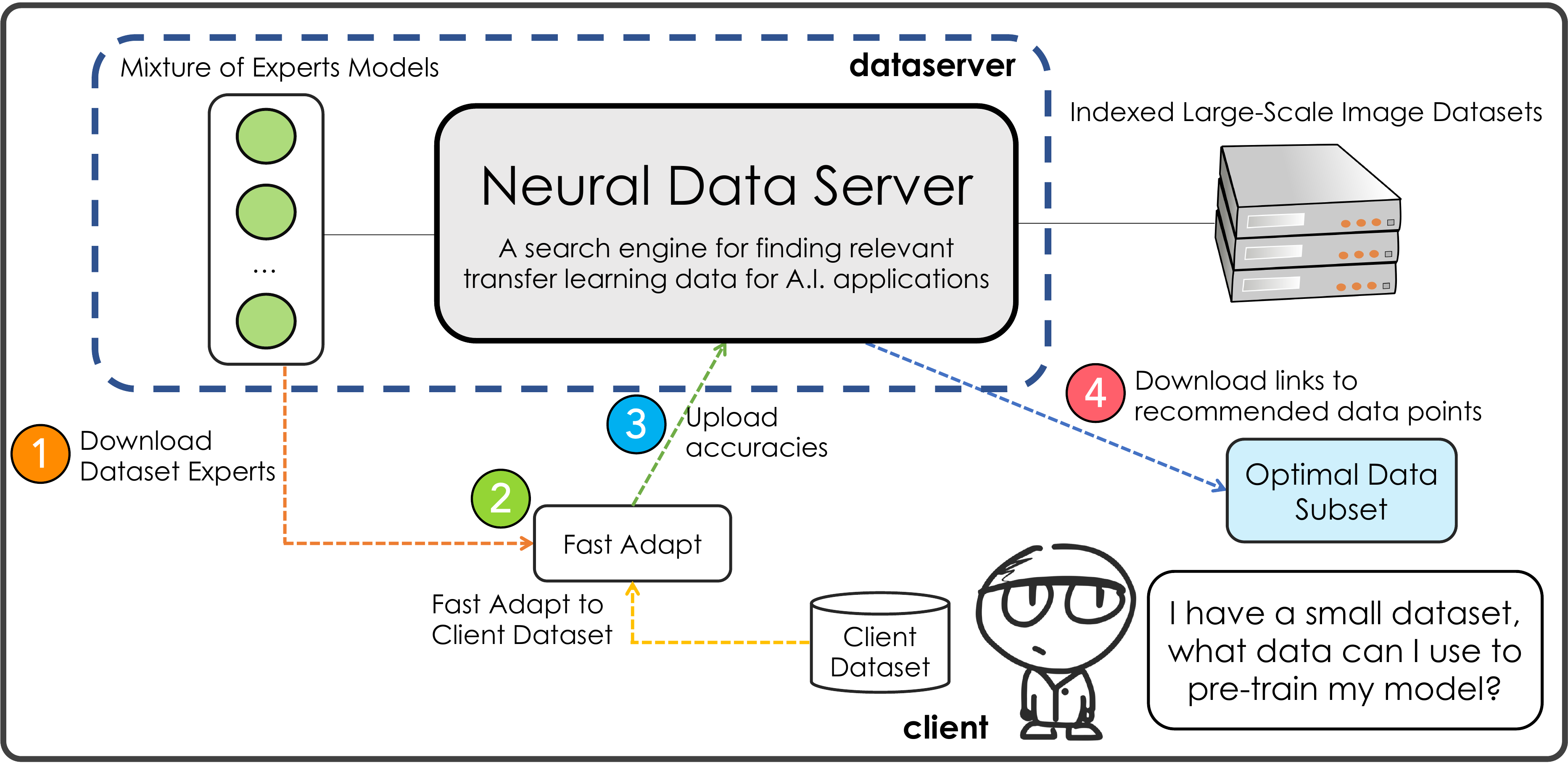}
\end{center}
\vspace{-6mm}
\caption{\footnotesize {\bf Neural Data Server}: Search engine for finding relevant transfer learning data for the user's target domain. In NDS, a {\ds} indexes several popular image datasets, represents them with a mixture-of-experts model, and uses {\cl}'s target data to determine most relevant samples. Note that NDS {\bf indexes} available public datasets and {\bf does not host} them. Data recommendation is done by providing {\bf links} to relevant examples. } 
\label{fig:NDS}
\vspace{-4mm}
\end{figure}

One of the critical   components of the new age of computer vision applications is the need for labeled data. To achieve high performance, typically a massive amount of data needs to be used to train deep learning models. Transfer learning provides a promising approach to reduce  the need for large-scale labeled data for each target application. In transfer learning, a neural network is pretrained~\cite{deeplab, He2017MaskR, Shelhamer2017FCN} on existing large generic datasets and then fine-tuned in the target domain. 
While transfer learning is a well studied concept that has been proven successful in many applications~\cite{deeplab, He2017MaskR, Shelhamer2017FCN}, deciding which data to use for pretraining the model is an open research question that has received surprisingly little attention in the literature. 
We argue that this is a crucial problem to be answered in light of the ever increasing scale of the available datasets. 

% To emphasize our point, the website of curated computer vision benchmarks\footnote{Website listing CV datasets: \url{https://www.visualdata.io/}} currently lists 367 public datasets, ranging from generic imagery, faces, fashion photos, to self-driving data. 
To emphasize our point, recent efforts on curating computer vision benchmarks\footnote{Websites listing CV datasets: {\scriptsize{\url{https://www.visualdata.io/}}}, {\scriptsize{\url{https://pytorch.org/docs/stable/torchvision/datasets.html}}}, {\scriptsize{\url{https://datasetsearch.research.google.com/}}}} list over 400 public datasets, ranging from generic imagery, faces, fashion photos, to self-driving data. Furthermore, the dataset sizes are significantly increasing: the recently released OpenImages ~\cite{OpenImages} contains 9M labeled images (600GB in size), and is 20 times larger compared to its predecessor MS-COCO ~\cite{Lin2014MicrosoftCC} (330K images, 30GB).
The video benchmark YouTube8m ~\cite{AbuElHaija2016YouTube8MAL} (1.9B frames, 1.5TB), is 800$\times$ larger compared to Davis ~\cite{Caelles2018The2D} (10k frames, 1.8GB), while the recently released autonomous driving dataset nuScenes~\cite{Caesar2019nuScenesAM} contains 100$\times$ the number of frames than KITTI~\cite{Geiger2012CVPR} which was released in 2012. 

\begin{figure*}[t]
\vspace{-1.5mm}
\includegraphics[width=\textwidth,height=60mm]{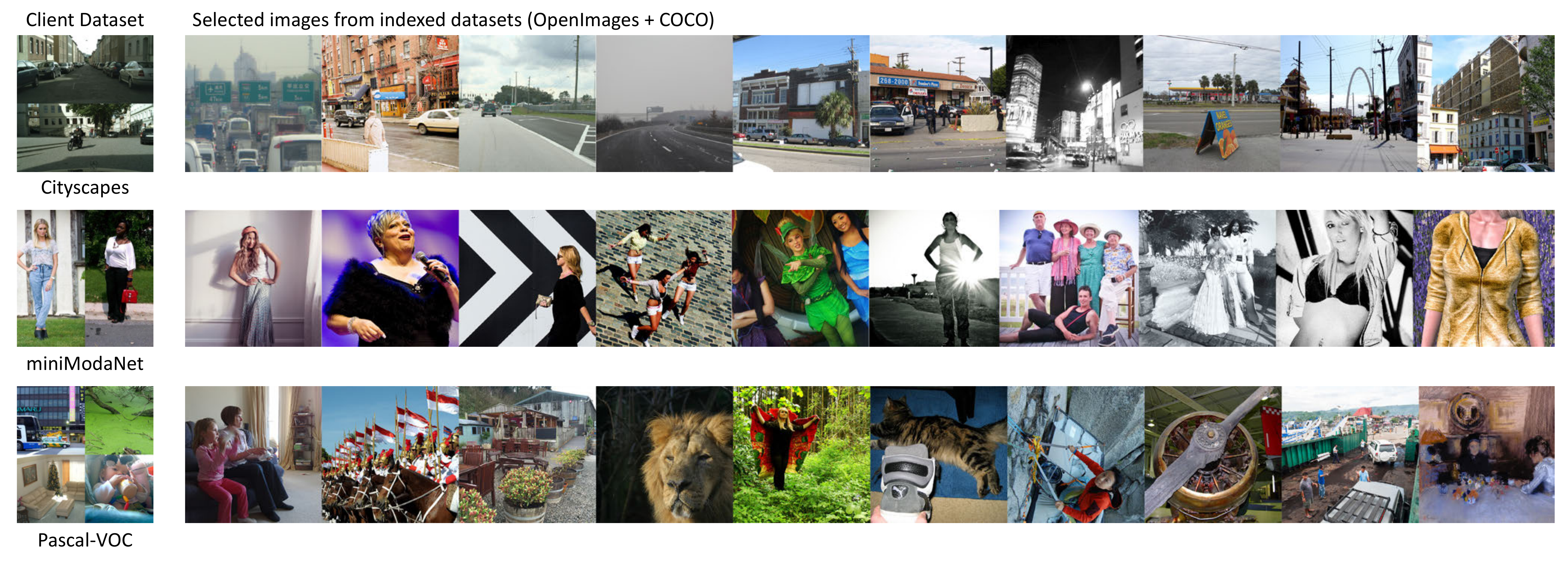}
\vspace{-8.5mm}
\caption{\footnotesize{Examples of images from the {\ds} (COCO+OpenImages) recommended to each {\cl} dataset by our Neural Data Server.}}
\label{example-visualize}
\vspace{-5mm}
\end{figure*}

It is evident that downloading and storing these datasets locally is already cumbersome and expensive.
This is further amplified by the computational resources required for training neural networks on this massive amount of data. The latter is an even more pronounced issue in research, where the network architectures are continuously being developed and possibly many need to be tested.  Furthermore, for commercial applications, data licensing may be another financial issue to consider. 
Recent works~\cite{he2018rethinking, ngiam2018domain} have also shown that there is not a ``the more the better" relationship between the amount of pretraining data and the downstream task performance. Instead, they showed that selecting an appropriate subset of the  data was important to achieve good performance on the target dataset. 

In this paper, we introduce Neural Data Server (NDS), a large-scale search engine for finding the most  
useful transfer learning data to the target domain. One can imagine NDS as a web-service where a centralized server, referred to as the {\ds}, recommends data to {\cl}s (Fig~\ref{fig:NDS}).  A {\cl} is an end-user with an A.I. application in mind, and has a small set of labeled target data. We assume that each client is only interested in downloading a subset of the server-indexed data that is most relevant to the client's target domain, limited to the user-specified budget (maximum desired size). 
We further require the transaction between the {\ds} and the {\cl} to be both computationally efficient and privacy-preserving. This means the client's data should not be visible to the server.
We also aim to minimize the amount of {\ds}'s online computation per client, as it may possibly serve  many clients in parallel. 

We index several popular image datasets and represent them using a mixture-of-experts (MoE) model, which we store on the {\ds}.  MoE is significantly smaller in size than the data itself, and is used to probe the usefulness of data in the {\cl}'s target domain. 
In particular, we determine the accuracy of each expert on the target dataset, and recommend data to the {\cl} based on these accuracies.

We experimentally show significant performance improvements on several downstream tasks and domains compared to baselines. 
Furthermore, we show that with only 20\% of  pretraining data,
 our method achieves comparable or better performance than pretraining on the entire {\ds}-indexed datasets. We obtain significant improvements over ImageNet pretraining by downloading only $26$ Gb of server's data in cases when training on the entire {\ds} ($538$ Gb) would take weeks. 
Our Neural Data Server will be made available as a web-service with the aim of improving performance and reducing the development cost of the end-users' A.I. applications.

\begin{figure*}[t]
\vspace{-2mm}
\begin{center}
\includegraphics[width=\textwidth]{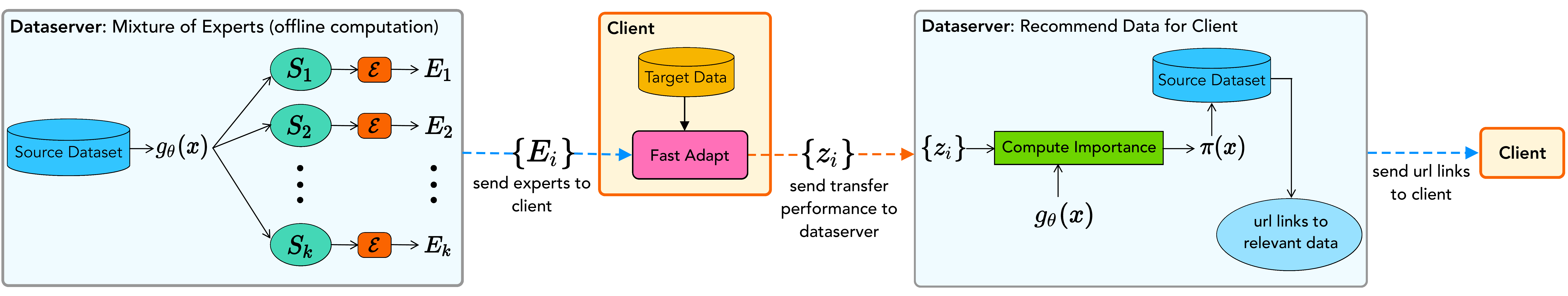}
\end{center}
\vspace{-7mm}
\caption{\footnotesize Overview of Neural Data Server. NDS consists of a {\ds} that represents indexed datasets using a mixture-of-experts model. Experts are sent to {\cl} in order to compute accuracies in the {\cl}'s target domain. These accuracies are then used by {\ds}  to recommend relevant data samples.}
\label{method-fig}
\vspace{-2.0mm}
\end{figure*}

\begin{figure*}[t]
\vspace{-4.5mm}
\begin{minipage}[t!]{0.48\textwidth}
\begin{algorithm}[H]
\begin{footnotesize}
  \caption{Dataserver's Modules}  
  \begin{algorithmic}[1]
  \State \textbf{Require} representation learning alg. $\mathcal{E}$, number of experts $K$
  \State $\vg_\theta \gets$ \textsc{HardGating($\gS, K$)} \Comment{{\tiny Sec~\ref{server}: partition $\gS$ into local subsets to obtain gating}}
  \State $\mathcal S_i:=\{\vx \in \gS | g_{\theta,i}(\vx) = 1\}$
  \State \textbf{procedure} \textsc{MoE}($\{\gS_i\}, \mathcal{E}, K$):
  \State \hskip2.0em \textbf{For} $i = 1, ..., K$
  \State \hskip4.0em Run $\mathcal{E}$ on $\mathcal S_i$ to obtain expert $e_{\theta_i}$
  \State \hskip2.0em \textbf{return} $\{e_{\theta_i}\}$
  \State \textbf{procedure} \textsc{OutputData}($\{S_i\}, \vz, \textnormal{budget}$):
  \State \hskip2.0em $\vw \gets \softmax{(\vz,T=0.1)}$
  \State \hskip2.0em $\pi(\vx) = \sum_{i=1}^{K} \ w_i \  g_{\theta,i}(\vx)   \frac{1}{|S_i|}$
  \State \hskip2.0em Sample $\gS_*$ from $\gS$ at rate according to $[\pi_{\vx_1},...,\pi_{\vx_n}]$ 
  \State \hskip2.0em \textbf{return} $\gS_*|_{\textnormal{budget}}$
  \end{algorithmic}
\end{footnotesize}
\end{algorithm}
\end{minipage}
\hfill
\begin{minipage}[t!]{0.48\textwidth}
	\begin{algorithm}[H]
	\begin{footnotesize}
  	\caption{Overview of our Neural Data Server}
  	\label{algo-overview}
  	\label{pseudocode}
  	\begin{algorithmic}[1]
  	\State \textbf{Input}: $\gS$ (source), $\gT$ (target), $b$ (desired budget of data)
  	\State \hskip2.0em $\{e_{\theta_i}\} \gets$ \textsc{MoE($\mathcal{D_S}, \mathcal{E}, K$)}
  	\State \hskip2.0em $\vz \gets$ \textsc{FastAdapt($\gT, \{e_{\theta_i}\}$)}
  	\State \hskip2.0em $\gS_* \gets$ \textsc{OutputData($\gS, \vz, b$)}
  	\State \hskip2.0em \textbf{return} $\gS_*$
  	\State \textbf{Output:} $\gS_* \in \gS$ to download
  	\end{algorithmic}
\end{footnotesize}
\end{algorithm}
\vspace*{-0.75cm}
\begin{algorithm}[H]
\begin{footnotesize}
  \caption{Client's Module} 
  \begin{algorithmic}[1]
  \State \textbf{procedure} \textsc{FastAdapt}($\mathcal{D_T}, \{e_{\theta_i}\}$):
  \State \hskip2.0em \textbf{For} $i = 1, ..., K$
  \State \hskip4.0em $\vz_i \gets$ \textsc{Performance/FineT}($e_{\theta_i}, \gT$) 
  \Comment{{\tiny Sec~\ref{server-client-trans}}}
  \State \hskip2.0em \textbf{return} $\vz$
  \end{algorithmic}
\end{footnotesize}
\end{algorithm}
\end{minipage}
\vspace{-3mm}
\end{figure*}

\vspace{-1.5mm}
\section{Related Work}
\vspace{-1.5mm}
\noindent\textbf{Transfer Learning.}
The success of deep learning and the difficulty of collecting large 
scale datasets has recently brought significant attention to the long existing history of transfer learning, cross-domain annotation and domain adaptation ~\cite{pan2009survey,csurka2017domain, acuna2018efficient, sun2017revisiting, Acuna_2019_CVPR,tremblay2018training}.
Specifically in the context of neural networks, fine-tuning a pretrained
model on a new dataset is the most common strategy for knowledge transfer.

Most literature in this domain analyzes the effect of pretraining on large-scale datasets~\cite{sun2017revisiting,mahajan2018exploring,imagenet_cvpr09} with respect to network architectures, network layers, and training tasks~\cite{Yosinski2014HowTA,Zamir2018TaskonomyDT}.
Concurrent with our work, Achille \etal~\cite{Achille2019Task2VecTE} proposes a 
%task embedding to identify 
framework for selecting
the best pre-trained feature extractor for a new task from a collection of classifiers. In contrast, our work aims to identify the optimal set of data points for pre-training.
Works most related to ours are~\cite{Cui2018LargeSF,ngiam2018domain} 
which show that pretraining on only relevant examples is important to achieve good performance on fine-grained classification tasks.
Specifically, in~\cite{Cui2018LargeSF} the authors use a predefined similarity metric between the source and target categories in order to greedily select the most similar categories from the source dataset to be used for pretraining.~\cite{ngiam2018domain}, on the other hand, exploits a model pretrained on the source domain to obtain pseudolabels of the target images, and uses these to re-weight the source examples.

 Unlike ours,~\cite{Cui2018LargeSF,ngiam2018domain} are limited to classification tasks, and do not easily scale to a constantly growing datacenter (the model needs to be retrained each time a new dataset is added). Thus, their approach does not naturally handle our scenario in which indexed datasets have diverse sets of tasks and labels, and where the number of indexed datasets may grow over time.

\noindent \textbf{Federated Learning.} ~\cite{McMahan2016FederatedLO, Bonawitz2017PracticalSA} introduce a distributed ML approach with the goal of training a centralized model on decentralized data over a large number of client devices, (\ie, mobile phones). Our work shares a similar  idea of restricting the visibility of data in a client-server model. However, in our case the representation of data is centralized (\ds) and the clients exploit the transfer learning scenario for their own (decentralized) models.

\noindent \textbf{Active and  Curriculum Learning.} In active learning~\cite{settles2009active} one searches over unlabeled data to find optimal samples to be labeled by an oracle, while in curriculum learning~\cite{bengio2009curriculum} subsets of data of increasing difficulty are sought for during training. In both scenarios, data search is performed at each iteration of training a particular model. Search is typically done by running inference on the data samples with the current snapshot of the model and selecting the examples based on uncertainty-based metrics. 
Our scenario differs in that we do not have the luxury of running inference with the {\cl}'s model on the massive amount of indexed data as this would induce a prohibitive computational overhead on the {\ds} per {\cl}. Moreover, we do not assume {{\ds} to have access to the {\cl}'s model: this would entail the {\cl}s to share their inference code which many users may not be willing to do.

\noindent \textbf{Learning to Generate Synthetic Images.}
Related to NDS are also \cite{ruiz2018learning,Metasim19,tripathi2019learning,mehta2019active}. These approaches aim 
to bridge the synthetic vs real imagery gap by optimizing/searching over the set of parameters of a surrogate function that interfaces with a synthesizer. 

In NDS, the search has to be done over massive (non-parametric) datasets and further, the target data cannot be sent to the server side. Our method is also significantly more computationally efficient.

\vspace{-1.5mm}
\section{Neural Data Server}
\vspace{-1.0mm}
Neural Data Server (NDS) is a search engine that aims to recommend transfer learning data. 
NDS consists of a {\ds} which has access to a massive source dataset(s), and aims to suggest most relevant data samples to a {\cl}. 
 A {\cl} is an end-user who wants a budget-constrained amount 
of
data to improve the performance of her/his model in the target domain in a transfer learning scenario. 
We note that the {\ds} does not host the data, and thus its recommendations are to be  provided as a list of urls to data samples hosted by the original datasets' providers. 

The {\ds}'s indexed datasets may or may not be completely labeled, and the types of labels (\eg, segmentation masks, detection boxes) across data samples may vary.  The {\cl}'s target dataset is considered to only have a small set of labeled examples, where further the type of labels may or may not be the same as the labels in the  {\ds}'s dataset(s).
The main challenge lies  in requiring the {\ds}-{\cl} transactions to have low computational overhead.
As in any search engine that serves information to possibly numerous users, we want the online computation performed by the {\ds} to be minimal. Thus we defer most of the computation to be performed on the {\cl}'s side, while still aiming for this process to be fast.
Furthermore, the transactions should ideally be privacy preserving for the {\cl}, \ie, the client's data nor the model's architecture are accessible, since the {\cl}  may have sensitive information such as hospital records or secret tech. 
In NDS, we represent the {\ds}'s data using a mixture-of-experts (MoE) trained on a self-supervised task. MoE naturally partition the indexed datasets into  different subsets and produce classifiers whose weights encode the representation of each of these subsets. The experts are trained offline and hosted on the {\ds} for online transactions with the clients. 
In particular, the experts are sent to each client and used as a proxy to determine the importance of {\ds}'s data samples for the {\cl}'s target domain.  

To compute importance, the experts are fast-adapted on the client's dataset, and their accuracy is computed on a simple self-supervised task. 
We experimentally validate that the accuracy of each adapted expert indicates the usefulness of the data partition used to train the expert. 
The {\ds} then uses these accuracies to construct the final list of data samples that are relevant for the {\cl}.  Figure~\ref{method-fig} provides an illustration while Algorithm~\ref{algo-overview} summarizes our NDS. 

In Section~\ref{problem-def} we formalize our problem. In Section~\ref{server} we describe how we train our mixture-of-experts model and analyze the different choices of representation learning algorithms for the experts ({\ds} side). In Section~\ref{server-client-trans} we propose how to exploit the experts' performance in the {\cl}'s target domain for data selection. 

\subsection{Problem Definition}
\label{problem-def}

Let  $\sX$ denote the input space (images in this paper), and $\sY_a$ a set of labels for a given task $a$. Generally, we will assume that multiple tasks are available, each associated with a different set of labels, and denote these by $\sY$. 
Consider also two different distributions over $\sX \times \sY$, called the source domain $\mathcal{D}_s$ and target domain 
$\mathcal{D}_t$. Let $\gS$ (\ds) and $\gT$ (\cl) be two sample sets drawn i.i.d from $\mathcal{D}_s$ and $\mathcal{D}_t$, respectively.

We assume that $|\gS|\gg|\gT|$. 

Our problem then relies on finding the subset $\gS_* \in \mathcal{P}(S)$, where $\mathcal{P}(S)$ is the power set of $\gS$, such that $\gS_* \cup \gT $ minimizes the risk of a model $h$ on the target domain:
\vspace{-2.0mm}
\begin{align}
  \label{eqn_intro}
\gS_* = \argmin_{\hat\gS \in \mathcal{P}(S) }\ \E_{(\vx, \hat \vy) \sim   \mathcal{D}_t} [ \mathcal{L}(h_{{\hat\gS \cup \gT}}(\vx), \hat \vy)]
\end{align}

\vspace{-2.0mm}
\noindent Here, $h_{{\hat\gS \cup \gT}}$ indicates that $h$ is trained on the union of data $\hat\gS$ and $\gT$.
Intuitively, we are trying to find the subset of data  from $\gS$ that helps to improve the performance of the model on the target dataset. % $\gT$.
However, what makes our problem particularly challenging and unique is that we are restricting the visibility of the data between the {\ds} and the {\cl}. 

This means that fetching the whole sample set $\gS$ is prohibitive for the client, as it is uploading its own dataset to the server. We tackle this problem by representing the {\ds}'s indexed dataset(s) with a set of classifiers that are agnostic of the client ({Section~\ref{server}}), and use these to optimize~\eqref{eqn_intro} on the \cl's side ({Section~\ref{server-client-trans}}). 

\subsection{Dataserver}
We now discuss our representation of the {\ds}'s indexed datasets. This representation is pre-computed offline and stored on the {\ds}.

\vspace{-3mm}
\label{server}
\subsubsection{Dataset Representation with Mixture-of-Experts}

We represent the {\ds}'s data $\mathcal S$ using the mixture-of-experts model~\cite{Jacobs1991AdaptiveMO}.
In MoE, one makes a prediction as:
\vspace{-3mm}
\begin{align}
\vy (\vx)= \sum_{i=1}^K g_{\theta,i}(\vx) e_{\theta_i}(\vx)
\end{align}
\vspace{-4mm}

Here,  $\vg_\theta$ denotes a gating function ($\sum_{i=1}^K g_{\theta,i}(.)=1$), $e_{\theta_i}$ denotes the  $i$-th expert model with learnable weights $\theta_i$, $\vx$ an input image, and $K$ corresponds to the number of experts. One can think of the gating function as softly assigning data points to each of the experts, which try to make the best guess on their assigned data points. 

The MoE model is trained by using maximum-likelihood estimation (MLE) on an objective  $\mathcal{L}$:
\vspace{-3mm}
\begin{align}
\mathbb\theta = \argmin_{\mathbb\theta} \E_{(\vx,\hat \vy)\sim \gS} [ \mathcal{L} (\vy(\vx),\hat \vy) ]
\end{align}

\vspace{-2.5mm}
We discuss the choices for the objective $\mathcal{L}$ in Sec~\ref{sec:experts}, dealing with the fact that the labels across the source datasets may be defined for different tasks.

While the MoE objective allows 
end-to-end training, the computational cost of doing so on a massive dataset is extremely high, particularly when $K$ is considerably large (we need to backpropagate gradients to every expert on every training example). A straightforward way to alleviate this issue is to associate each expert with a local cluster defined by a hard gating, as in~\cite{Hinton2015DistillingTK, gross2017hard}.  In practice, we define a gating function $g$ that partitions the dataset into mutually exclusive subsets $\mathcal S_i$, and train one expert per subset. This makes training easy to parallelize as each expert is trained independently on its subset of data. Furthermore, this allows for new datasets to be easily added to the {\ds} by training additional experts on them, and adding these to {\ds}. This avoids re-training MoE over the full indexed set of datasets. 

In our work, we use two simple partitioning schemes to determine the gating: (1) superclass partition, and (2) unsupervised partition. For superclass partition (1), we represent each class $c$ in the source dataset as the mean of the image features $f_c$ for category $c$, and perform $k$-means clustering over $\{f_c\}$. This gives a partitioning where each cluster is a superclass containing a subset of similar categories. This partitioning scheme only applies to datasets with class supervision.
For unsupervised partitioning (2), we partition the source dataset  using $k$-means clustering on the image features. In both cases, the image features are obtained from a pretrained neural network (\ie, features extracted from the penultimate layer of a network pre-trained on ImageNet). 

\vspace{-3mm}
\subsubsection{Training the Experts}
\label{sec:experts}

 We discuss two different scenarios to train the experts. In the simplified scenario, the tasks defined for both the {\ds}'s and {\cl}'s datasets are the same, \eg, classification. In this case, we simply train a classifier for the task for each subset of the data in $\mathcal S$. We next discuss a more challenging case where the tasks across datasets differ. 

Ideally, we would like to learn a representation that can generalize to a variety of downstream tasks and can therefore be used 
in a task agnostic fashion. 
To this end, we  use a self-supervised method to train the MoE. 
In self-supervision, one leverages a simple surrogate task that can be used  to learn a meaningful representation. 

Furthermore, this does not
require any labels to train the experts which means that the {\ds}'s dataset may or may not be labeled beforehand. 
This is useful if the client desires to obtain raw data and label the relevant subset on its own. 
To be specific, we select classifying  image rotation as the task for self-supervision as in~\cite{gidaris2018unsupervised}, which
showed this to be a simple yet powerful proxy for representation learning.
Formally, given an image $\vx$, we define its corresponding self-supervised label $\vy$  by performing 
 a set of geometric transformations $\{r(\vx, j)\}_{j=0}^{3}$ on $\vx$,  where $r$ is an image rotation operator, and $j$ defines a particular rotation by one of the predefined angles, $\{0, 90, 180, 270\}$.
We then minimize the following learning objective for the experts:

\vspace{-3mm}
\begin{align}
\gL(\theta_i) = -\sum_{\vx\in \mathcal S_i} \sum_{j=0}^{3} \log e_{{\theta_i}}(r(\vx,j))_j
\end{align}
\vspace{-4.5mm}

\noindent Here, index $j$ in $e(.)_j$ denotes the output value for class $j$. 

\subsection{Dataserver-Client Transactions}
In this section, we describe the transactions between the {\ds} and {\cl} that determines the relevant subset of the server's data. The {\cl} first downloads the experts  in order to measure their performance on the {\cl}'s dataset. 
If the tasks are similar, 
we perform a quick adaptation of the experts on the {\cl}'s side. 
Otherwise, we evaluate the performance of the experts on the {\cl}'s data using the surrogate task (i.e image rotation) (Section~\ref{server-client-trans}).
The performance of each expert is sent back to the {\ds}, which uses this information as a proxy to  determine which data points are relevant to the {\cl} (Section~\ref{selection}). We describe these steps in more detail in the following subsections. 

\vspace{-3mm}
\subsubsection{\textsc{FastAdapt} to a Target Dataset (on Client)}
\label{server-client-trans}

\vspace{-1mm}
\paragraph{Single Task on Server and Client:} We first discuss the case where the dataset task is the same for both the {\cl} and the {\ds}, e.g., classification. While the task may be the same, the label set may not be (classes may differ across domains). 
An intuitive way to adapt the experts is to remove their classification head that was trained on the server, and learn a small decoder network on top of the experts's penultimate representations on the client's dataset, as in~\cite{Zamir2018TaskonomyDT}. 
For classification tasks, we learn a simple linear layer on top of each pre-trained expert's representation for a few epochs. We then evaluate the target's task performance on a held-out validation set using the adapted experts. We denote the accuracy for each adapted expert $\hat e_{\theta_i}$ as $z_i$.  

\vspace{-3mm}
\paragraph{Diverse Tasks on Server and Client:}  To generalize to unseen tasks and be further able to handle cases where the labels are not available on the {\cl}'s side,  we propose to evaluate the performance of the common self-supervised task used to train the experts on the {\ds}'s data. Intuitively, if the expert performs well on the self-supervised task on the target dataset, then the data it was trained on is likely relevant for the {\cl}. Specifically, we use the self-supervised experts trained to learn image rotation, and evaluate the proxy task performance (accuracy) of predicting image rotation angles on the target images:
\vspace{-3mm}
\begin{align}
z_i = \frac{1}{4 |\mathcal{\gT}|} \sum_{\vx \in \gT } \sum_{j=0}^{3}  \mathbbm{1}\big(\argmax_{k} [e_{{\theta_i}} (\vr(\vx,j))_k] ==j\big)
\end{align}

\vspace{-3.0mm}
\noindent Here, index $k$ in $e(.)_k$ denotes the output value for class $k$. 

Note that in this case we do not adapt the experts on the target dataset (we only perform inference).

\vspace{-2mm}
\subsubsection{Data Selection (on Dataserver)}
\label{selection}

We now aim to assign a weighting to each of the data points in the source domain $\gS$ to reflect how well the source data contributed to the transfer learning performance. The accuracies $\vz$ from the client's \textsc{FastAdapt} step  are normalized to $[0, 1]$ and fed into a \textit{softmax} function with temperature $T=0.1$. These are then used as importance weights $ w_i$ for estimating how relevant is the representation learned by a particular expert for the target task's performance. We leverage this information to weigh the individual data points $\vx$. More specifically, each source data $\vx$ is assigned a probabilistic weighting: 
\vspace{-3mm}
\begin{align}
\pi(\vx) = \sum_{i=1}^{K} \ w_i \  g_{\theta,i}(\vx)   \frac{1}{|S_i|}
\end{align}
\vspace{-3.5mm}

\noindent Here, $|S_i|$ represents the size of the subset that an expert $e_{\theta_i}$ was trained on. 
Intuitively, we are weighting the set of images associated to the $i$-th expert  and uniformly sampling from it.
We construct our dataset by sampling examples from $\mathcal{\gS}$ at a rate according to $\bm{\pi}={[\pi_{\vx_1},\pi_{\vx_2},\hdots,\pi_{\vx_n}]}^T$. 

\subsection{Relation to Domain Adaptation}
 If we assume that the client and server  tasks are the same then our problem can be interpreted as domain adaptation in each of the subset $\hat\gS \in \mathcal{P}(S)$ and the following generalization bound from ~\cite{BenDavid2009ATO} can be used:
\vspace{-2.5mm}
\begin{align}
    \label{eqn_bound}
\varepsilon_{\mathcal{T}}(h) < \varepsilon_{\hat{\mathcal{S}}}(h)+ \frac{1}{2} d_{\mathcal{H} \Delta \mathcal{H} }(\hat{\mathcal{S}}, \mathcal{T})
\end{align} 

\vspace{-2.5mm}

\noindent where   $\varepsilon$ represents the risk of a hypothesis function $h \in \mathcal{H}$ and  $d_{\mathcal{H} \Delta \mathcal{H} }$  is the ${\mathcal{H} \Delta \mathcal{H} }$ divergence ~\cite{BenDavid2009ATO}, which relies on the capacity of $\mathcal{H}$ to distinguish between data points from $\hat\gS$ and $\gT$, respectively.

Let us further assume that the risk of the hypothesis function $h$ on any subset $\hat\gS$ is similar such that: $\varepsilon_{\hat{\mathcal{S}}}(h) \approx \varepsilon(h) \ \ \forall  \hat\gS \in \mathcal{P}(\gS)
$.
Under this assumption,  minimizing equation~\ref{eqn_intro} is equivalent to finding the subset $\gS_*$ that minimizes the divergence with respect to $\gT$. Formally,
\vspace{-2.0mm}
\begin{align}
\gS_*=\argmin_{\hat\gS } d_{\mathcal{H} \Delta \mathcal{H} (\hat\gS,\gT) }
\end{align} 
\vspace{-3.5mm}

\noindent In practice, it is  hard to compute $d_{\mathcal{H} \Delta \mathcal{H} }$ 
 and this is often approximated by a \emph{proxy $\mathcal{A}$-distance} ~\cite{NIPS2006_2983,chen2015marginalizing,Ganin2015DomainAdversarialTO}.
 A classifier that discriminates between the two domains and whose risk $\varepsilon$ is used to approximate the second part of the equation~\ref{eqn_bound}. 
 \vspace{-2.0mm}
\begin{align}
\hat{d}_{\mathcal{H}} \approx \hat{d}_{\mathcal{A}} \approx  2(1-2\varepsilon)
\end{align}

\vspace{-2.0mm}
\noindent Note that doing so would require having access to $\gS$ and $\gT$ in at least one of the two sides (i.e to train the new discriminative classifier) and this is prohibitive in our scenario. 
In our case, we compute the domain confusion between $\hat\gS$ and $\gT$ by evaluating the performance of expert $e_i$ on the target domain. We argue that this proxy task performance (or error rate) is an appropriate proxy distance that serves the same purpose but does not violate the data visibility condition. Intuitively, if the features learned on the subset cannot be discriminated from features on the target domain, the domain confusion is maximized. We empirically show the correlation between the domain classifier 
and our proposed proxy task performance in our experiments. 

\vspace{-0mm}
\section{Experiments}
\vspace{-1mm}

\begin{table*}[t]
\centering
\addtolength{\tabcolsep}{-2pt}
\resizebox{\textwidth}{!}{
\footnotesize
\begin{tabular}{c|c|c|c|c|c||c|c|c||c|c|c}
\toprule
\multicolumn{3}{c|}{Pretrain Server Data (COCO + OpenImages)} & \multicolumn{8}{c}{Client Dataset} \\ 
\hline \hline
\multicolumn{2}{c|}{Sampled Data Size} & \multirow{2}{*}{Method} & \multicolumn{3}{c||}{PASCAL-VOC2007} & \multicolumn{3}{c||}{miniModaNet} & \multicolumn{3}{c}{Cityscapes} \\ \cline{0-1}
File Size & \# Images & & $AP^{bb}$ & $AP^{bb}_{50}$ & $AP^{bb}_{75}$ & $AP^{bb}$ & $AP^{bb}_{50}$ & $AP^{bb}_{75}$ & $AP^{bb}$ & $AP^{bb}_{50}$ & $AP^{bb}_{75}$ \\ \hline \hline
\multicolumn{3}{c|}{ImageNet Initialization} & 44.30 & 73.66 & 46.44 & 33.40 & 57.98 & 35.00 & 34.94 & 59.86 & 35.69 \\
\hline \hline
\multirow{2}{*}{26GB / 538GB} & \multirow{2}{*}{90K (5\%)} & Uniform Sampling & 47.61 & 76.88 & 51.95 & 35.64 & 58.40 & 39.09 & 36.49 & 61.88 & 36.36 \\ 
 &  & \bf{NDS} & \bf{48.36} & \bf{76.91} & \bf{52.53} & \bf{38.84} & \bf{61.23} & \bf{43.86} & \bf{38.46} & \bf{63.79} & \bf{39.59} \\ \hline \hline
\multirow{2}{*}{54GB / 538GB} & \multirow{2}{*}{180K (10\%)} & Uniform Sampling & 48.05 & 77.17 & 52.04 & 35.78 & 58.50 & 39.71 & 36.41 & 61.22 & 37.17 \\ 
 &  & \bf{NDS} & \bf{50.28} & \bf{78.61} & \bf{55.47} & \bf{38.97} & \bf{61.32} & \bf{42.93} & \bf{40.07} & \bf{65.85} & \bf{41.14} \\
\bottomrule
\end{tabular}
}
\vspace{-3mm}
\caption{\small{Results for object detection on the 3 client datasets. Scores are measured in \%. }\label{combined-server-bbox}}
\vspace{-1mm}
\end{table*}

\begin{table}[ht]
\vspace{-1mm}
\small
\resizebox{\linewidth}{!}{%
\addtolength{\tabcolsep}{-2pt}
\begin{tabular}{c|c|cc|cc}
\toprule
Data (\# Images) & Method & $AP^{bb}$ & $AP^{bb}_{50}$ & $AP$ & $AP_{50}$ \\
\hline \hline
0 & ImageNet Initial. & 36.2 & 62.3 & 32.0 & 57.6 \\
\hline \hline
\multirow{2}{*}{23K} & Uniform Sampling & 38.1 & 64.9 & 34.3 & 60.0 \\
 & \textbf{NDS} & 40.7 & 66.0 & 36.1 & 61.0 \\
\hline \hline
\multirow{2}{*}{47K} & Uniform Sampling & 39.8 & 65.5 & 34.4 & 60.0 \\
 & \textbf{NDS} & \bf{42.2} & \bf{68.1} & \bf{36.7} & \bf{62.3} \\
\hline \hline
\multirow{2}{*}{59K} & Uniform Sampling & 39.5 & 64.9 & 34.9 & 60.4 \\
 & \textbf{NDS} & 41.7 & 66.6 & \bf{36.7} & 61.9 \\
\hline \hline
118K & Full COCO & 41.8 & 66.5 & 36.5 & \bf{62.3} \\
\bottomrule
\end{tabular}%
}
\vspace{-3mm}
\caption{\footnotesize{Transfer learning results for instance segmentation with Mask R-CNN on Cityscapes by selecting images from COCO.}}
\vspace{-1mm}
\label{maskrcnn-seg}
\end{table}

\begin{table}[ht]
\vspace{-1mm}
\small
\resizebox{\linewidth}{!}{%
\addtolength{\tabcolsep}{-2pt}
\begin{tabular}{c|c|cc|cc}
\toprule
Data (\# Images) & Method & $AP^{bb}$ & $AP^{bb}_{50}$ & $AP$ & $AP_{50}$ \\
\hline \hline
 0 & ImageNet Initial. & 36.2 & 62.3 & 32.0 & 57.6 \\
\hline \hline
\multirow{2}{*}{118K} & Uniform Sampling & 37.5 & 62.5 & 32.8 & 57.2 \\
 & \textbf{NDS} & \bf{39.9} & \bf{65.1} & \bf{35.1} & \bf{59.8}  \\
\hline \hline
\multirow{2}{*}{200K} & Uniform Sampling & 37.8 & 63.1 & 32.9 & 57.8 \\
 & \textbf{NDS} & \bf{40.7} & \bf{65.8} & \bf{36.1} & \bf{61.2} \\
\bottomrule
\end{tabular}%
}
\vspace{-3mm}
\caption{\footnotesize{Transfer learning results for instance segmentation with Mask R-CNN on Cityscapes by selecting images from OpenImages.}}
\label{maskrcnn-seg-openimages}
\vspace{-3mm}
\end{table}

\begin{figure*}[ht]
\vspace{-5mm}
\begin{minipage}{.428\linewidth}
\begin{table}[H]
\resizebox{\linewidth}{!}{%
\addtolength{\tabcolsep}{-4pt}
\begin{tabular}{|ll|c|c|c|c|c|}
\hline
\multicolumn{2}{|c|}{\multirow{2}{*}{\textbf{Pretrain. Sel. Method}}} & \multicolumn{5}{c|}{\textbf{Target Dataset}} \\
\multicolumn{2}{|c|}{} & Stanf. Dogs & Stanf. Cars & Oxford-IIIT Pets & Flowers 102 & CUB200 Birds \\
\hline
0\% & Random Init. & 23.66 & 18.60 & 32.35 & 48.02 & 25.06 \\
\hline
100\% & Entire Dataset & 64.66 & 52.92 & 79.12 & 84.14 & 56.99 \\
\hline
\multirow{4}{*}{20\%} & Uniform Sample & 52.84 & 42.26 & 71.11 & 79.87 & 48.62 \\
 & NDS (SP+TS) & 72.21 & 44.40 & 81.41 & 81.75 & 54.00 \\
 & NDS (SP+SS) & 73.46 & 44.53 & 82.04 & 81.62 & 54.75 \\
 & NDS (UP+SS) & 66.97 & 44.15 & 79.20 & 80.74 & 52.66 \\
\hline
\multirow{4}{*}{40\%} & Uniform Sample & 59.43 & 47.18 & 75.96 & 82.58 & 52.74 \\
 & NDS (SP+TS) & 68.66 & 50.67 & 80.76 & 83.31 & 58.84 \\
 & NDS (SP+SS) & 69.97 & 51.40 & 81.52 & 83.27 & 57.25 \\
 & NDS (UP+SS) & 67.16 & 49.52 & 79.69 & 83.51 & 57.44 \\ 
\hline
\end{tabular}%
}
\vspace{-3mm}
\caption{ Ablation experiments on gating and expert training. SP=Superclass Partition, UP=Unsupervised Partition, TS=Task-Specific experts (experts trained on classif. labels), and SS=Self-Supervised experts (experts trained to predict image rotation).}
\label{classification-result}
\end{table}
\end{minipage}
\hspace{1.5mm}
\begin{minipage}{.266\linewidth}
\begin{figure}[H]
\includegraphics[width=\linewidth]{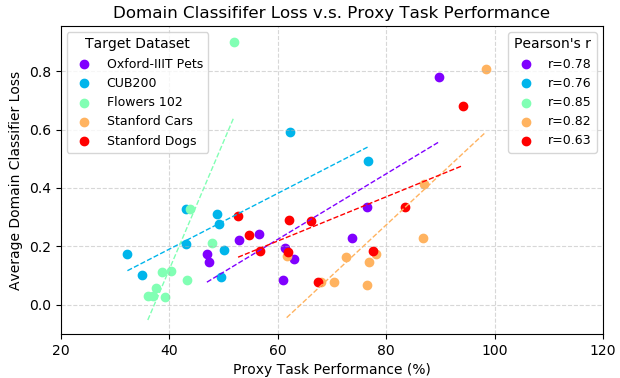}

\vspace{-3mm}
\caption{\footnotesize{Relationship between domain classifier and proxy task performance on subsets $\hat\gS$.}}
\label{figure-domain-confusion}
\end{figure}
\end{minipage}
\hspace{1.5mm}
\begin{minipage}{.272\linewidth}
\begin{table}[H]
\resizebox{\linewidth}{!}{%
\addtolength{\tabcolsep}{-4pt}
\begin{tabular}{|l|l|c|c|}
\hline
Data & Method & Oxford-IIIT Pet & CUB200 Birds \\
\hline
\multirow{4}{*}{20\%} & Uniform Samp. & 71.1 & 48.6 \\
 & KNN + ~\cite{Cui2018LargeSF} & 74.4 & 51.6 \\
 & ~\cite{ngiam2018domain} & 81.3 & 54.3 \\
 & NDS & \bf{82.0} & \bf{54.8} \\
 \hline
\multirow{4}{*}{40\%} & Uniform Samp. & 76.0 & 52.7 \\
 & KNN + ~\cite{Cui2018LargeSF} & 78.1 & 56.1 \\
 & ~\cite{ngiam2018domain} & 81.0 & \bf{57.4} \\
 & NDS & \bf{81.5} & 57.3 \\
 \hline
\multicolumn{2}{|c|}{Entire ImageNet} & 79.1 & 57.0 \\
\hline
\end{tabular}%
}
\vspace{-2mm}
\caption{\footnotesize{Transfer learning performance on classification datasets comparing data selection methods. \label{baseline-comparisons}}}
\end{table}
\end{minipage}
\end{figure*}

\begin{figure*}[ht]
\vspace{-1mm}
\includegraphics[width=\linewidth,height=15.5mm]{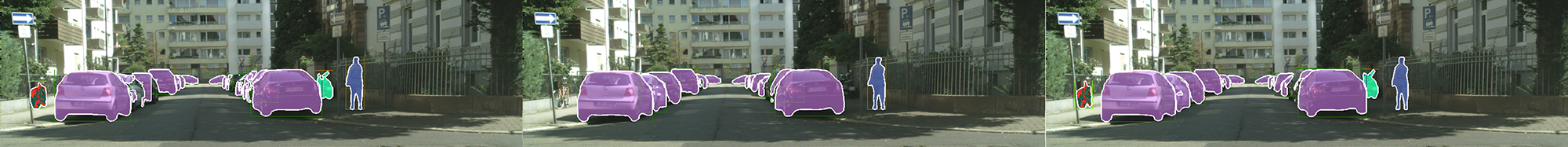}
\includegraphics[width=\linewidth,height=15.5mm]{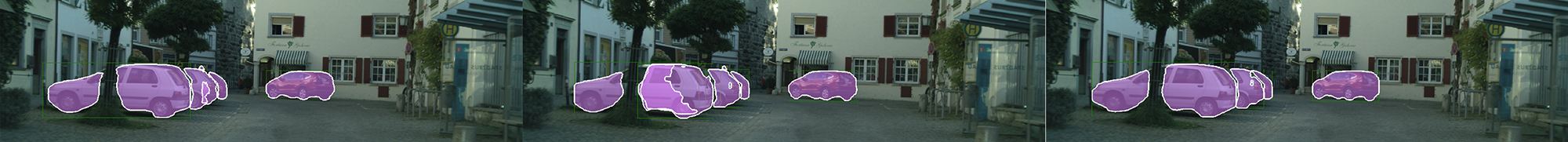}
\vspace{-7mm}
\caption{\footnotesize{Instance segmentation results on Cityscapes using network pre-trained from ImageNet initialization (\textbf{left}), 47K images uniformly sampled (\textbf{middle}), and 47K images from NDS (\textbf{right}). Notice that the output segmentations generally look cleaner when training on NDS-recommended data.}}
\label{cityscapes-visualize}. 
\vspace{-5mm}
\end{figure*}

\begin{figure*}[ht]
\includegraphics[width=\linewidth,height=36mm]{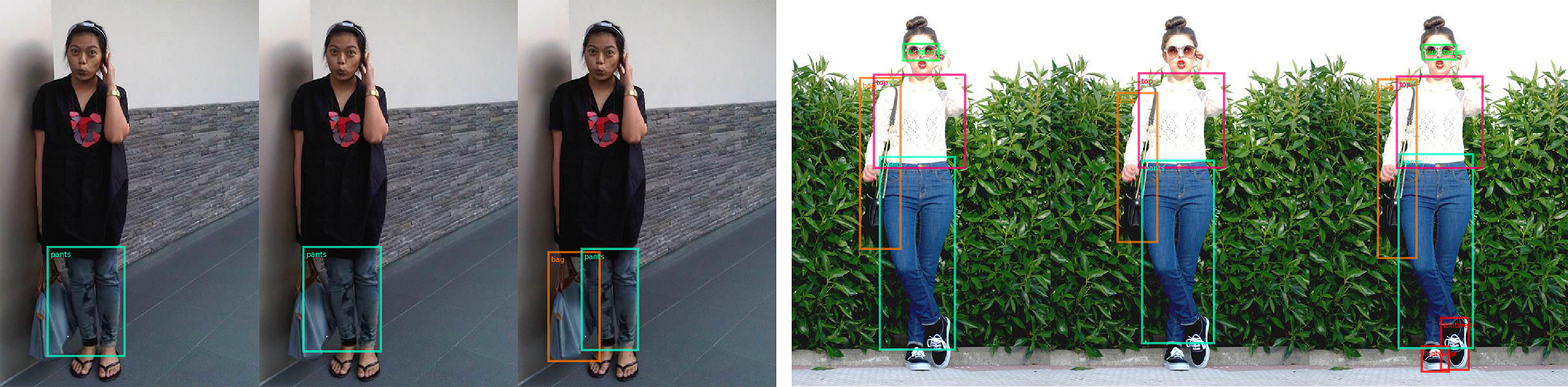}
\vspace{-6.5mm}
\caption{\footnotesize{Object detection results in miniModaNet using network pre-trained from ImageNet initialization (\textbf{left}), 90K images uniformly sampled (\textbf{middle}), and 90K images sampled using NDS (\textbf{right}). A score threshold of 0.6 is used to display these images.} \label{fig:modanet-visualize}}
\vspace{-2mm}
\end{figure*}
We perform experiments in the tasks of  classification, detection, and instance segmentation. We experiment with 3 datasets on the sever side and 7 on the client side.

\subsection{Support for Diverse Clients and Tasks}
In this section, 
we provide an extensive evaluation of our approach on three different client's scenarios:  autonomous driving, fashion and general scenes. 
In each of them, the client's goal is to improve the performance of its downstream task (\ie, object detection or instance segmentation) by pretraining in a budget-constrained amount of data. 
Here, the {\ds} is the same and indexes the massive OpenImages~\cite{OpenImages} and MS-COCO ~\cite{Lin2014MicrosoftCC} datasets.
Specifically, our server dataset can be seen as the union of COCO and OpenImages ~\cite{OpenImages, Lin2014MicrosoftCC} (approx $538$ GB)
represented in the weights of the self-supervised trained experts ($2$ GB).

\noindent \textbf{Autonomous Driving:}
Here, we use  Cityscapes~\cite{Cordts2016TheCD} as the client's dataset, which contains $5000$ finely annotated images divided into $2975$ training and $500$ validation images. %, and $1525$ test 
Eight
object classes are provided with per-instance annotation.
In practice, this simulates the scenario of a client that wants to crunch its performance numbers
by pretraining on some data.
This scenario is ubiquitous among state-of-the-art instance and semantic segmentation approaches on the Cityscapes leaderboard ~\cite{Takikawa2019GatedSCNNGS,He2017MaskR,Zhu2018ImprovingSS}.

\noindent \textbf{Fashion:}
We use the ModaNet dataset~\cite{Zheng2018ModaNetAL} to simulate a client that wants to improve its models' performance in the task of object detection of fashion related objects.
ModaNet is a large-scale street fashion dataset consisting of $13$ classes of objects and $55,176$ annotated images.
Since the effectiveness of pre-training diminishes with the size of the dataset~\cite{he2018rethinking}, we create a small version of the dataset for our experiments. 
This constitutes of $1000$ training  and $1000$ validation images that are randomly selected but keeping the same class distribution of the original dataset. 
We call it miniModaNet in our experiments. 

\noindent \textbf{General Scenes:} We use PASCAL VOC object detection ~\cite{Everingham2009ThePV} as the client's dataset for this scenario.
The task in this case is object detection on $20$ object classes. We use the \texttt{trainval2007} set containing $5011$ images for training and evaluate on \texttt{test2007} containing $4962$ images. 

\noindent \textbf{Evaluation:}
We use Intersection-Over-Union (IoU) to measure
client's performance in its downstream task. 
Specifically, we follow the MS-COCO evaluation style and compute IoU at three different thresholds: a) $0.50$, b) $0.75$, c) an average of ten thresholds  $(.5:.05:.95)$. 
The same evaluation style is used for both, object detection and instance segmentation. Notice however that in the case of instance segmentation, the overlap is based on segmented regions. 

\noindent \textbf{Baselines:} 
In this regime, we compare our approach vs no pretraining, uniform sampling, and pretraining on the whole server dataset (\ie, MS-COCO). In all cases, we initialize with ImageNet pretrained weights as they are widely available and this has become a common practice.

\noindent \textbf{Implementation Details:}
\textbf{Client.} We use Mask-RCNN~\cite{He2017MaskR} with a ResNet50-FRN backbone detection head as the client's network. After obtaining a subset of $\gS$, the client pre-trains a network on the selected subset and uses the pre-trained model as initialization for fine-tuning using the client (target) dataset. For object detection, we pre-train with a 681 class (80 class from COCO, 601 class from OpenImages) detection head using bounding box labels. For instance segmentation, we pre-train with 80 class (for COCO) or 350 class (for OpenImages) detection head using object mask labels.

\textbf{Server.} For all self-supervised experts, we use \textit{ResNet18} ~\cite{He2015DeepRL},
and train our models to predict image rotations.  MS-COCO and OpenImages  are partitioned into $ K = 6$ and $K=50$ experts, respectively. 

\vspace{-4mm}
\subsubsection{Qualitative and Quantitative Results}
\vspace{-2mm}
\noindent \textbf{Object Detection:} Table~\ref{combined-server-bbox} reports the average precision at various IoU of the client's network pre-trained using data selected using different budgets and methods. First, we see that a general trend of pre-training the network on sampled detection data helps performance when fine-tuning on smaller client detection datasets compared to fine-tuning the network from ImageNet initialization. By pre-training on 90K images from COCO+OpenImages, we observe a 1-5\% gain in AP at 0.5 IoU
across all 3 client (target) datasets. This result is consistent with~\cite{Li2019AnAO} which suggests that a pre-training task other than classification is beneficial for improving transfer performance on localization tasks. Next, we see that under the same budget of 90K/180K images from the server, pre-training with data selected by NDS outperforms the baseline which uses images randomly sampled from $\gS$ for all client datasets.

\noindent \textbf{Instance Segmentation:} Table~\ref{maskrcnn-seg} reports the instance segmentation performance by sampling 23K, 47K, and 59K images from COCO for pre-training on Cityscapes. We can see that pre-training using subsets selected by NDS is 2-3\% better than the uniform sampling baseline. Furthermore, using 40\% (47K/118K), or 50\% (59K/118K) images from COCO yields comparable (or better) performance to using the entire 100\% (118K) of data. Table~\ref{maskrcnn-seg-openimages} shows the results of sampling 118K, 200K images from OpenImages dataset as our server dataset.

\noindent\textbf{Qualitative Results:} Figure~\ref{fig:modanet-visualize} shows qualitative results on \textsc{miniModaNet} from detectors pre-trained from Imagenet, uniformly sampled images from $\gS$, and images sampled using NDS. In the cases shown, the network pre-trained using the data recommended by NDS shows better localization ability, and is able to make more accurate predictions.

\subsection{Support for Diverse Clients Same Task}
For completeness, and in order to compare to stronger baselines that are limited to classification tasks, we also quantitatively evaluate  the performance of NDS in the same-client-same-task regime. 
In this case, the task is set to be classification and the server indexes the Downsampled ImageNet ~\cite{chrabaszcz2017downsampled} dataset. This  a variant of ImageNet ~\cite{imagenet_cvpr09} resized to 32$\times$32.
In this case, we use $K=10$ experts. % #to be send to a client.   

\noindent \textbf{Client's Datasets:} We experiment with several small classification datasets.
Specifically, we use Stanford Dogs~\cite{KhoslaYaoJayadevaprakashFeiFei_FGVC2011}, %(120class/12000train/8580val)
Stanford Cars~\cite{KrauseStarkDengFei-Fei_3DRR2013}, %(196class/8144train/8041val)
Oxford-IIIT Pets~\cite{parkhi12a}, %(37class/3680train/3369val)
Flowers 102~\cite{Nilsback08}, %(102class/2040train/6149val)
and CUB200 Birds~\cite{WahCUB_200_2011} %(200class/5994train/5794val)
 as client datasets.

\noindent \textbf {Implementation Details:}
We use ResNet18~\cite{He2015DeepRL} as our client's network architecture, and an input size of $32 \times 32$ during training. Once subsets of server data are selected, we pre-train on the selected subset and evaluate the performance by fine-tuning on the client (target) datasets. 

\noindent\textbf{Comparison to data selection methods:} Cui \etal~\cite{Cui2018LargeSF} and Ngiam \etal~\cite{ngiam2018domain} recently proposed data selection methods for improving transfer learning for classification tasks.
In this restricted regime, we can compare to these methods. 
Specifically, we compare our NDS with~\cite{ngiam2018domain}, where they sample data based on the probability over source dataset classes computed by pseudo-labeling the target dataset with a classifier trained on the source dataset. We also create a baseline KNN by adapting Cui \etal 's method~\cite{Cui2018LargeSF}. Here, we sample from the most similar categories measured by the mean features of categories between the client and server data. 
We emphasize that the previous two approaches are limited to the classification task, and cannot handle diverse tasks. Furthermore, they do not scale to datasets beyond classification, and~\cite{ngiam2018domain} does not scale to a growing {\ds}. Our approach achieves comparable results to~\cite{ngiam2018domain}, and can be additionally applied to source datasets with no classification labels such as MS-COCO, or even datasets which are not labeled. 

\begin{figure}[t!]
\centering
\vspace{-5mm}
\begin{minipage}{0.49\linewidth}
\includegraphics[width=1\linewidth, trim=20 0 20 0,clip]{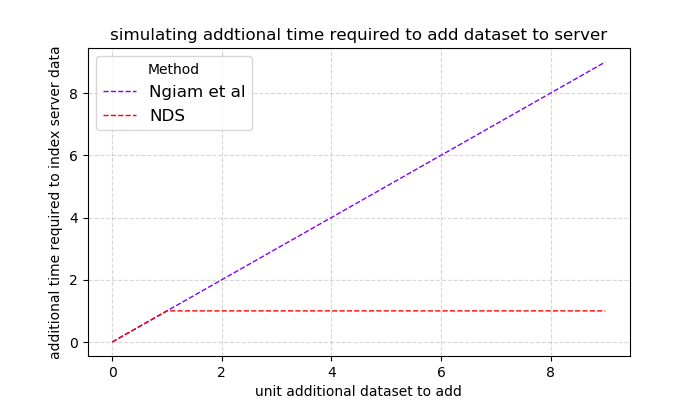}
\end{minipage}
\begin{minipage}{0.49\linewidth}
\vspace{3mm}
\caption{\footnotesize Simulating  an incrementally growing {\ds}, and the time required to ``train" a model to represent the server. We NDS compare to the baseline of~\cite{ngiam2018domain} (which is limited to classification tasks).}
\label{fig:scalability}
\end{minipage}
\vspace{-5mm}
\end{figure}

\subsection{Ablation Experiments}
\vspace{-1mm}

\noindent \textbf{Domain Confusion:} To see how well the performance of the proxy task reflects the domain confusion, we perform an experiment comparing the proxy task performance and $\hat{d}_{\mathcal{A}}(\hat\gS, \gT)$. 
To estimate $\hat{d}_{\mathcal{A}}$, we follow the same idea from  \cite{NIPS2006_2983,chen2015marginalizing,Ganin2015DomainAdversarialTO} 
 and for each subset $\hat\gS$, we estimate the domain confusion.
 Figure \ref{figure-domain-confusion} shows the domain confusion vs the proxy task performance using several classification datasets as the target (client) domain. 
 In this plot,
 the highest average loss corresponds to the subset with the highest domain confusion (\ie, $\gS_i$ that is the most indistinguishable from the target domain).
 Notice that this correlates with the expert that gives the highest proxy task performance.

\noindent\textbf{Ablation on gating and expert training:} In Table \ref{classification-result}, we compare different instantiations of our approach on five client classification datasets. For all instantiations, pre-training on our selected subset significantly outperforms the pre-training on a randomly selected subset of the same size. Our result in Table \ref{classification-result} shows that under the same superclass partition, the subsets obtained through sampling according to the transferability measured by self-supervised experts (SP+SS) yield a similar downstream performance compared to sampling according to the transferability measured by the task-specific experts (SP+TS). This suggests that self-supervised training for the experts can successfully be used as a proxy to decide which data points from the source dataset are most useful for the target dataset. 

\noindent\textbf{Scalability:} 
% Fig~\ref{fig:scalability} analyzes the (simulated) required training time of the server as a new dataset is being incrementally added to it.  We simulate a comparison between~\cite{ngiam2018domain} (which needs to retrain the model each time a dataset is added, and thus scales linearly) and NDS. 
Fig~\ref{fig:scalability} analyzes the (simulated) required training time of the server as a new dataset is being incrementally added to it.  We simulate a comparison between~\cite{ngiam2018domain} (which needs to retrain the model on all datasets each time a dataset is added, and thus scales linearly) and NDS (where expert training is only ran on the additional dataset).

\begin{figure*}[htb!]
\begin{center}
\addtolength{\tabcolsep}{-5pt}
\begin{tabular}{cccc}
\includegraphics[height=2.62cm,trim=160 0 150 0, clip]{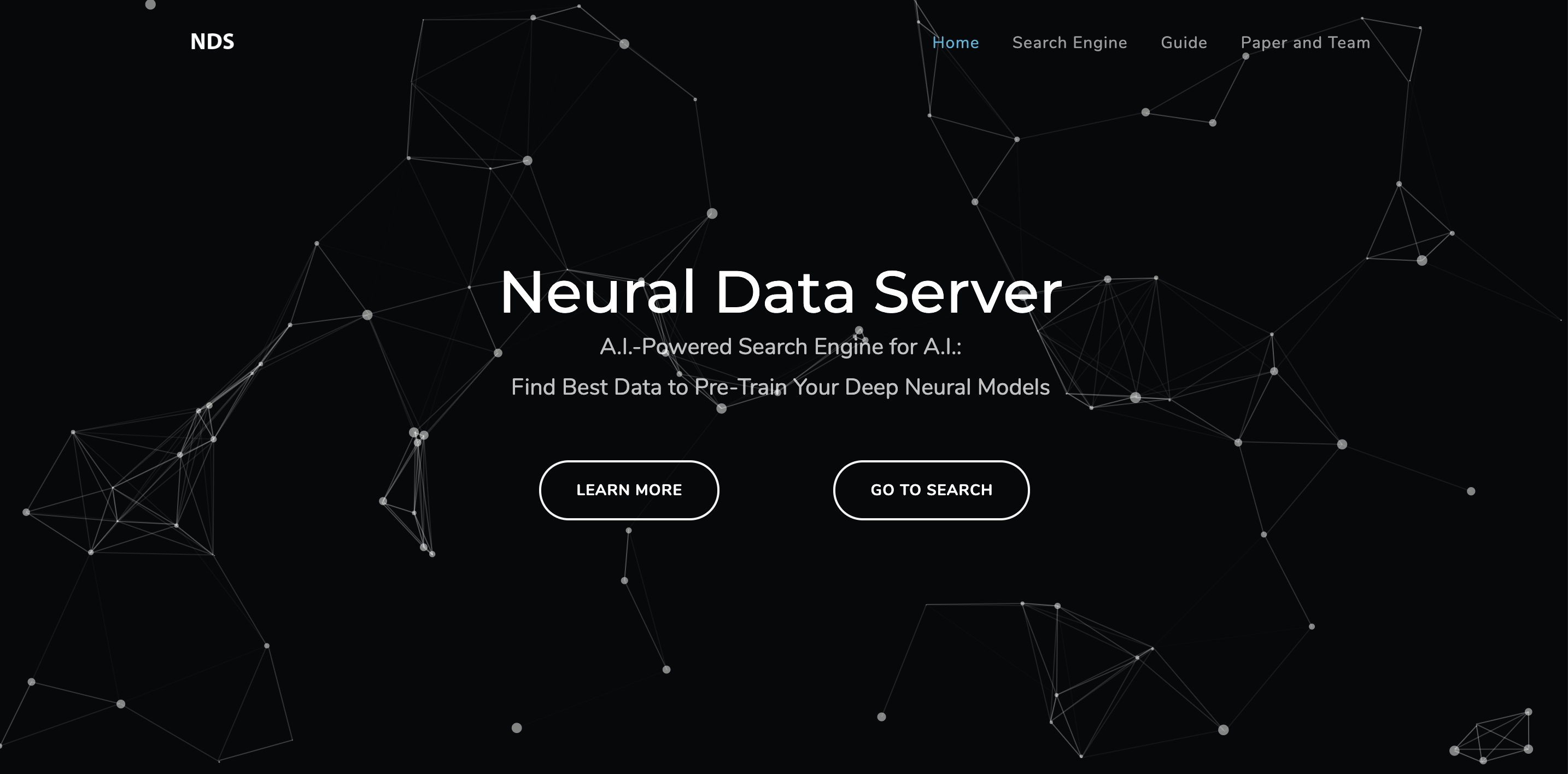} &
\includegraphics[height=2.62cm,trim=0 0 00 0, clip]{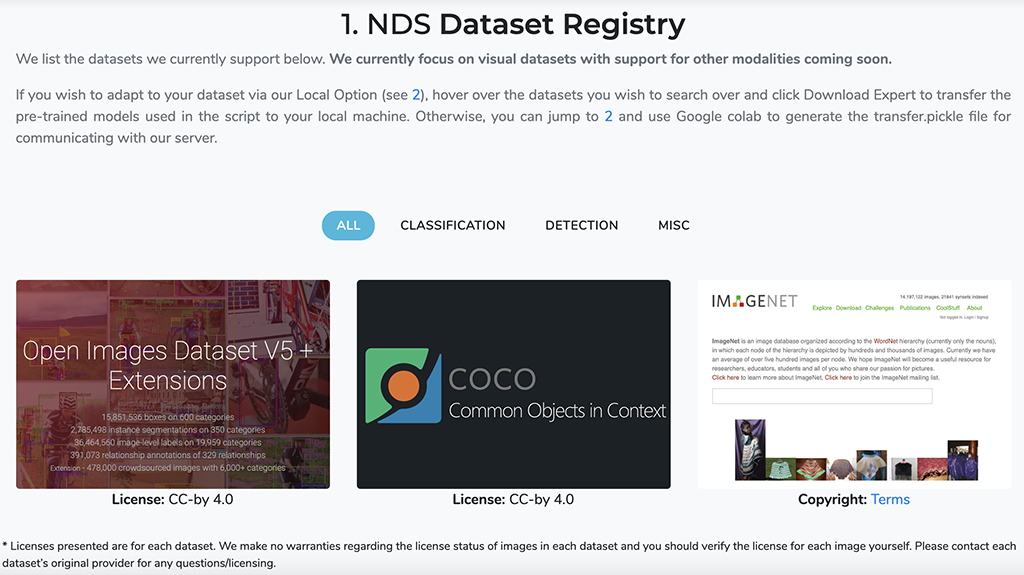} &
\includegraphics[height=2.62cm,trim=60 0 60 0, clip]{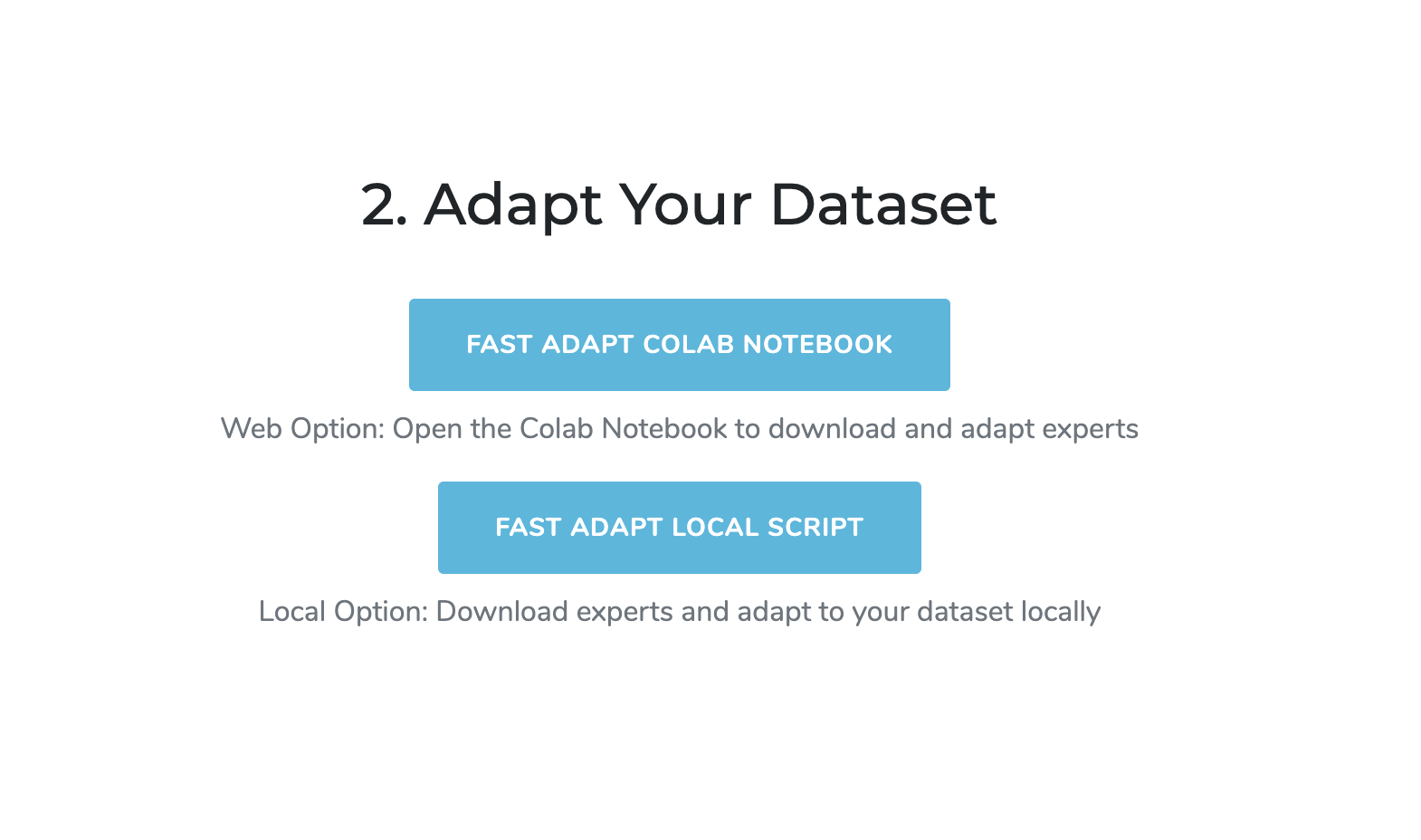}&
\includegraphics[height=2.62cm,trim=00 0 10 0, clip]{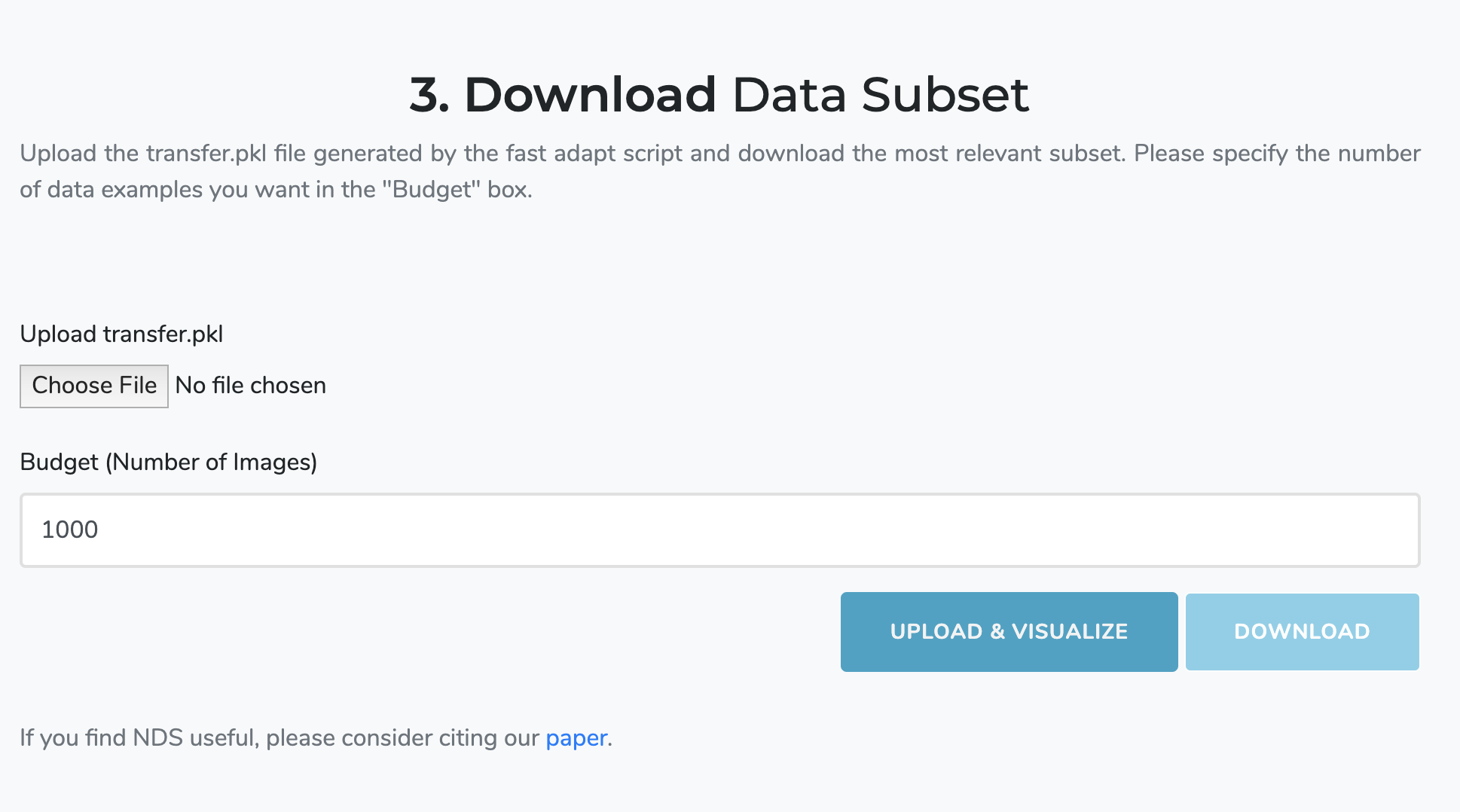}\\
{\href{http://aidemos.cs.toronto.edu/nds/}{\footnotesize\color{magenta}{aidemos.cs.toronto.edu/nds/}}} & \small{Dataset Registry} & \small{Adapt Experts} & \small{Download Recommended Data}
\end{tabular}
\end{center}
\vspace{-5mm}
\caption{\small Our Neural Data Server  web-service. Note that NDS does not host datasets, but rather links to datasets hosted by original providers.}
\label{fig:web-interface}
\vspace{-1mm}
\end{figure*}

\begin{table*}[t]
\footnotesize
\resizebox*{\linewidth}{!}{
\addtolength{\tabcolsep}{-1pt}
\begin{tabular}{c|c|c|c|c}
\toprule
Dataset & Images & Class & Task & Evaluation Metric \\
\hline \hline
Downsampled ImageNet~\cite{chrabaszcz2017downsampled} & 1281167 & 1000 & classification & - \\
OpenImages~\cite{OpenImages} & 1743042 & 601(bbox) / 300(mask) & detection & - \\
COCO~\cite{Lin2014MicrosoftCC} & 118287 & 80 & detection & - \\
\hline \hline
VOC2007~\cite{Everingham2009ThePV} & 5011(trainval) / 4962(test) & 20 & detection & mAP \\
miniModaNet~\cite{Zheng2018ModaNetAL} & 1000(train) / 1000(val) & 13 & detection & mAP \\
Cityscapes~\cite{Cordts2016TheCD} & 2975(train) / 500(val) & 8 & detection & mAP \\
\hline \hline
Stanford Dogs ~\cite{KhoslaYaoJayadevaprakashFeiFei_FGVC2011} & 12000(train) / 8580(val) & 120 & classification & Top-1 \\
Stanford Cars ~\cite{KrauseStarkDengFei-Fei_3DRR2013} & 8144(train) / 8041 (val) & 196 & classification & Top-1 \\
Oxford-IIIT Pets ~\cite{parkhi12a} & 3680(train) / 3369(val) & 37 & classifiation & Top-1 \\
Flowers 102 ~\cite{Nilsback08} & 2040(train) / 6149(val) & 102 & classification & Top-1 \\
CUB200 Birds ~\cite{WahCUB_200_2011} & 5994(train) / 5794(val) & 200 & classification & Top-1 \\
\bottomrule
\end{tabular}%
}
\vspace{-3mm}
\caption{\small Summary of the number of images, categories, and evaluation metrics for datasets used in our experiments. We used 10 datasets (3 server datasets and 7 client datasets) to evaluate NDS.}
\vspace{-2mm}
\label{dataset-stats}
\end{table*}

\begin{table*}[t]
\vspace{0mm}
\small
\resizebox{\linewidth}{!}{%
\addtolength{\tabcolsep}{-1pt}
\begin{tabular}{c|c|cc|cc|cccccccc}
\toprule
Data (\# Images) & Method & $AP^{bb}$ & $AP^{bb}_{50}$ & $AP$ & $AP_{50}$ & car & truck & rider & bicycle & person & bus & mcycle & train \\
\hline \hline
0 & ImageNet Initialization & 36.2 & 62.3 & 32.0 & 57.6 & 49.9 & 30.8 & 23.2 & 17.1 & 30.0 & 52.4 & 17.9 & 35.2 \\
\hline \hline
\multirow{2}{*}{23K} & Uniform Sampling & 38.1 & 64.9 & 34.3 & 60.0 & 50.0 & 34.2 & 24.7 & 19.4 & 32.8 & 52.0 & 18.9 & 42.1\\
 & \textbf{NDS} & 40.7 & 66.0 & 36.1 & 61.0 & 51.3 & 35.4 & 25.9 & 20.4 & 33.9 & 56.9 & 20.8 & 44.0 \\
\hline \hline
\multirow{2}{*}{47K} & Uniform Sampling & 39.8 & 65.5 & 34.4 & 60.0 & 50.7 & 31.8 & 25.4 & 18.3 & 33.3 & 55.2 & 21.2 & 38.9 \\
 & \textbf{NDS} & 42.2 & 68.1 & 36.7 & 62.3 & 51.8 & 36.9 & 26.4 & 19.8 & 33.8 & 59.2& 22.1 & 44.0 \\
\hline \hline
\multirow{2}{*}{59K} & Uniform Sampling & 39.5 & 64.9 & 34.9 & 60.4 & 50.8 & 34.8 & 26.3 & 18.9 & 33.2 & 55.5  & 20.8 & 38.7 \\
 & \textbf{NDS} & 41.7 & 66.6 & 36.7 & 61.9 & 51.7 & 37.2 & 26.9 & 19.6 & 34.2 & 56.7 & 22.5 & 44.5 \\
\hline \hline
118K & Full COCO & 41.8 & 66.5 & 36.5 & 62.3 & 51.5 & 37.2 & 26.6 & 20.0 & 34.0 & 56.0 & 22.3 & 44.2 \\
\bottomrule
\end{tabular}%
}
\vspace{-3mm}
\caption{\small Transfer to instance segmentation with Mask R-CNN~\cite{He2017MaskR} on Cityscapes by selecting images from COCO.}
\label{table:maskrcnn-seg-coco}

\vspace{2mm}
\small
\resizebox{\linewidth}{!}{%
\addtolength{\tabcolsep}{-1pt}
\begin{tabular}{c|c|cc|cc|cccccccc}
\toprule
Data (\# Images) & Method & $AP^{bb}$ & $AP^{bb}_{50}$ & $AP$ & $AP_{50}$ & car & truck & rider & bicycle & person & bus & mcycle & train \\
\hline \hline
 0 & ImageNet Initialization & 36.2 & 62.3 & 32.0 & 57.6 & 49.9 & 30.8 & 23.2 & 17.1 & 30.0 & 52.4 & 17.9 & 35.2 \\
\hline \hline
\multirow{2}{*}{118K} & Uniform Sampling & 37.5 & 62.5 & 32.8 & 57.2 & 49.6 & 33.2 & 23.3 & 18.0 & 30.8 & 52.9 & 17.4 & 37.1 \\
 & \textbf{NDS} & 39.9 & 65.1 & 35.1 & 59.8 & 51.6 & 36.7 & 24.2 & 18.3 & 32.4 & 56.4 & 18.0 & 42.8 \\
\hline \hline
\multirow{2}{*}{200K} & Uniform Sampling & 37.8 & 63.1 & 32.9 & 57.8 & 49.7 & 31.7 & 23.8 & 17.8 & 31.0 & 51.8 & 18.4 & 38.8 \\
 & \textbf{NDS} & 40.7 & 65.8 & 36.1 & 61.2 & 51.4 & 38.2 & 24.2 & 17.9 & 32.3 & 57.8 & 19.7 & 47.3 \\
\bottomrule
\end{tabular}%
}
\vspace{-3mm}
\caption{\small Transfer to instance segmentation with Mask R-CNN~\cite{He2017MaskR} on Cityscapes by selecting images from OpenImages.}
\label{table:maskrcnn-seg-openimages}
\end{table*}

\noindent \textbf{Limitations and Discussion:} A limitation in our method is that the annotation quality/statistics in the {\ds} datasets is not considered. This is shown in our instance segmentation experiment where the gains from pre-training on images sampled from OpenImages is smaller than pre-training on MS-COCO. This is likely due to the fact that MS-COCO has on average $\sim$7 instance annotations per image while OpenImages contains many images without mask annotations or at most $\sim$2 instance annotations per image. OpenImages has further been labeled semi-automatically and thus in many cases the annotations are noisy.

\vspace{-2mm}
\section{Conclusion}
\vspace{-1.5mm}
In this work, we propose a novel method that aims to optimally select subsets of data from a large dataserver given a particular target client. 
In particular, we represent the server's data with a mixture of experts 
trained on a simple self-supervised task.
These are then used as a proxy to determine the most important subset of the data that the server should send to the client.
We experimentally show that our method is general and can be applied to any pre-training and fine-tuning scheme and that our approach even handles the case where no labeled data is available (only raw data). We hope that our work opens a more effective way of performing transfer learning in the era of massive datasets. 
In the future, we aim to increase the capability of NDS to also support other modalities such as 3D, text and speech.

\vspace{-3.5mm}
\paragraph{Acknowledgments:} The authors acknowledge partial support by NSERC. SF acknowledges the Canada CIFAR AI
Chair award at Vector Institute.  We thanks Relu Patrascu for his continuous infrastructure support. We also thank Amlan Kar, Huan Ling and Jun Gao for early discussions, and Tianshi Cao and Jonah Philion for feedback in the manuscript.}

%%%%%%%%% BODY TEXT

\begin{figure*}[h]
\begin{center}
\includegraphics[width=.246\linewidth]{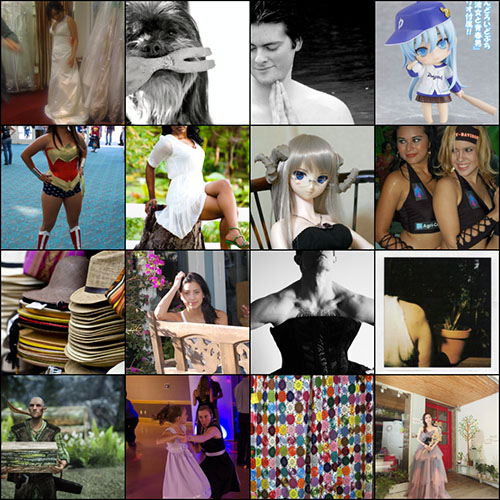}
\vspace{0.1mm}
\includegraphics[width=.246\linewidth]{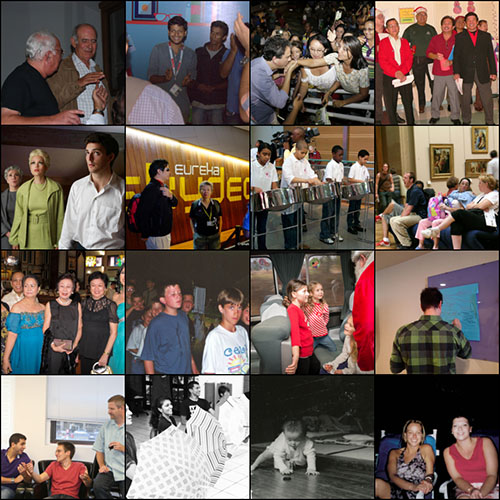}
\vspace{0.1mm}
\includegraphics[width=.246\linewidth]{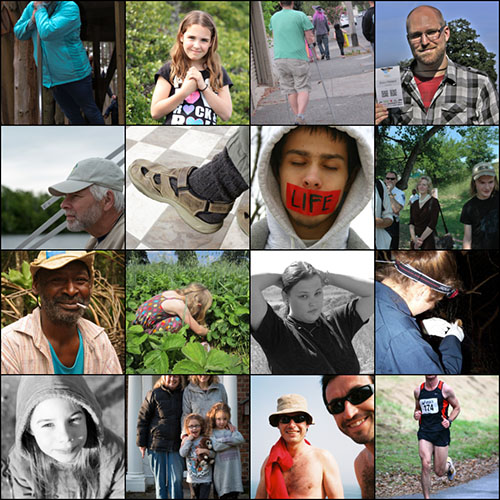}
\vspace{0.1mm}
\includegraphics[width=.246\linewidth]{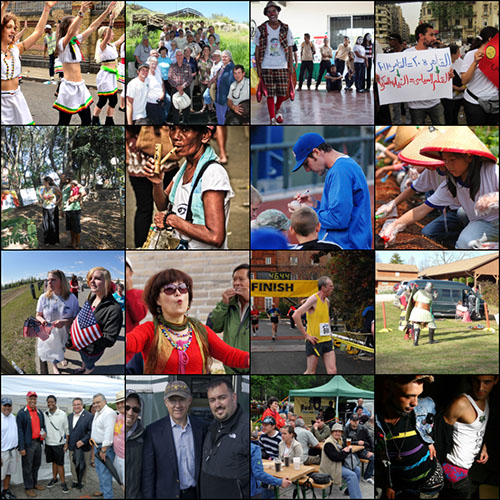}
\vspace{0.1mm}
\includegraphics[width=.246\linewidth]{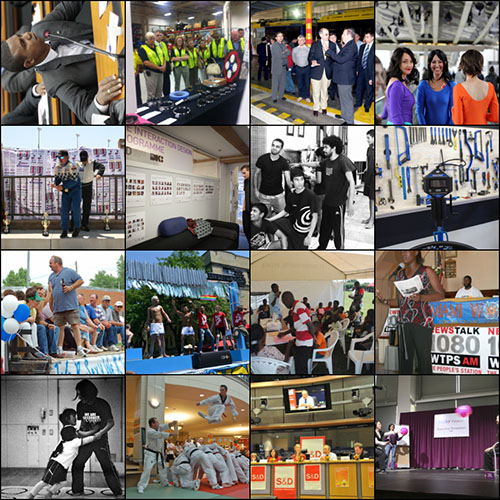}
\includegraphics[width=.246\linewidth]{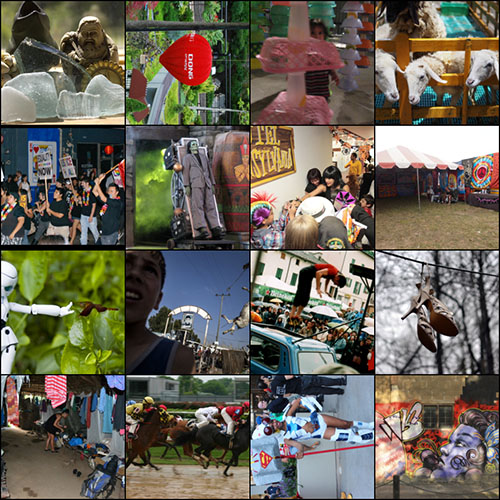}
\includegraphics[width=.246\linewidth]{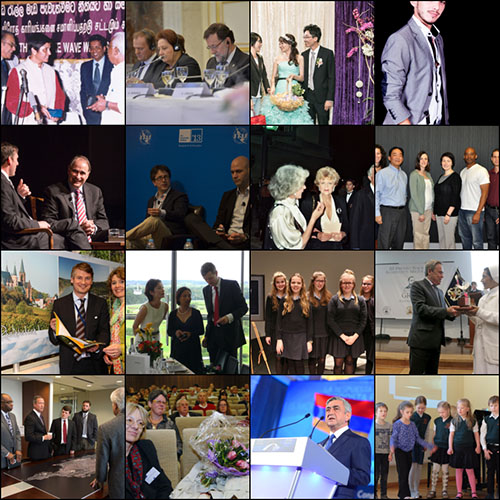}
\includegraphics[width=.246\linewidth]{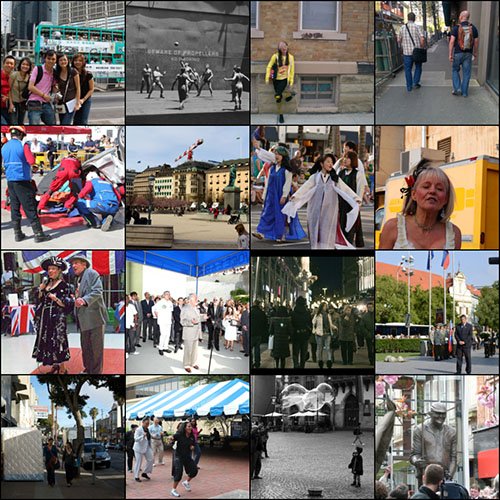}
\end{center}
\vspace{-5mm}
\caption{Top 8 clusters from COCO+OpenImages corresponding to the best performing expert adapted on miniModaNet.}
\label{fig:modanet-clusters}
\end{figure*}

\begin{figure*}[h]
\begin{center}
\includegraphics[width=.246\linewidth]{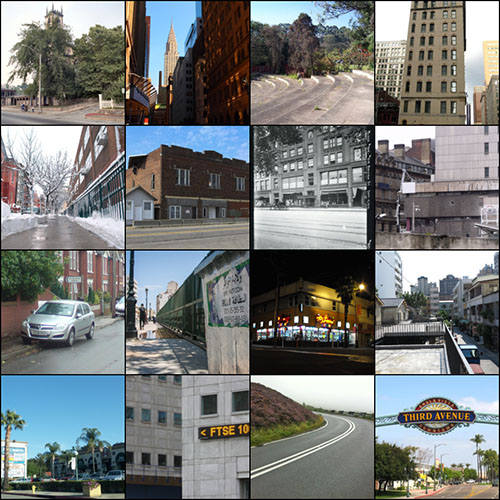}
\vspace{0.1mm}
\includegraphics[width=.246\linewidth]{figs/openimages-cluster/43.jpg}
\vspace{0.1mm}
\includegraphics[width=.246\linewidth]{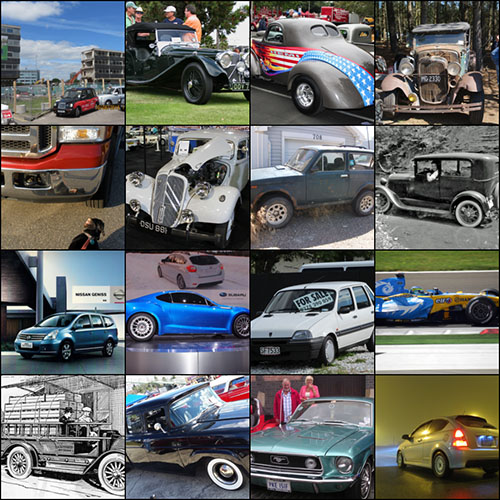}
\vspace{0.1mm}
\includegraphics[width=.246\linewidth]{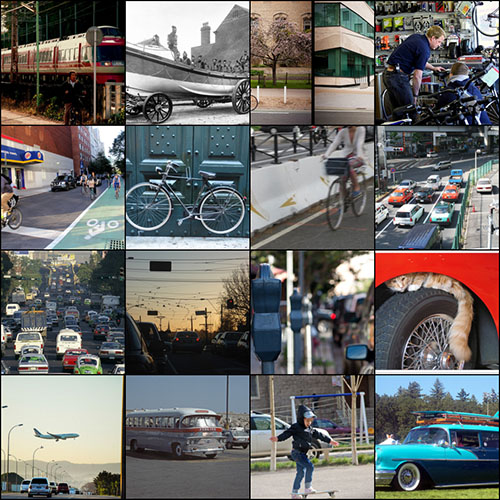}
\vspace{0.1mm}
\includegraphics[width=.246\linewidth]{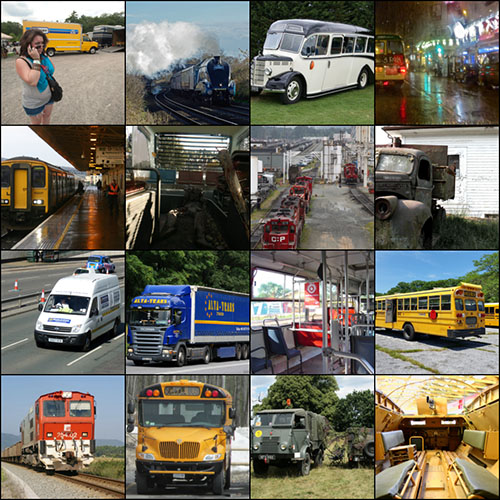}
\includegraphics[width=.246\linewidth]{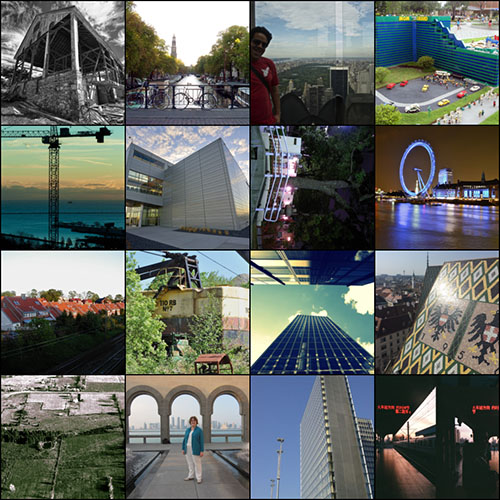}
\includegraphics[width=.246\linewidth]{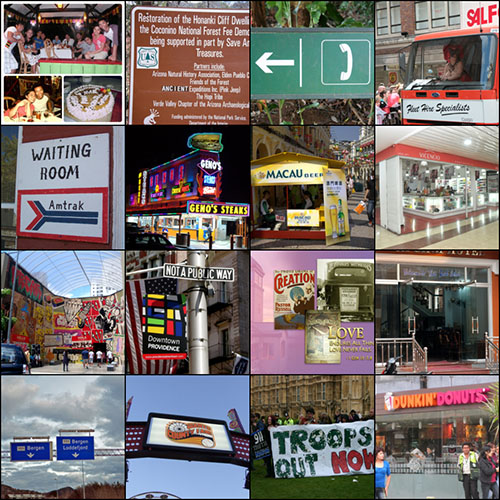}
\includegraphics[width=.246\linewidth]{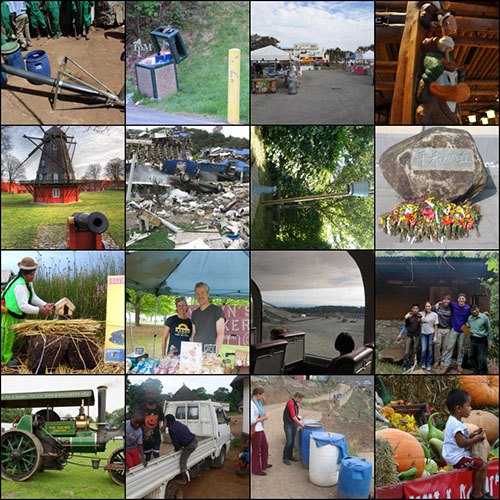}
\end{center}
\vspace{-5mm}
\caption{Top 8 clusters from COCO+OpenImages corresponding to the best performing expert adapted on Cityscapes.}
\label{fig:cityscapes-clusters}
\end{figure*}

\begin{figure*}[t]
\begin{center}
\includegraphics[width=.246\linewidth]{figs/openimages-cluster/18.jpg}
\vspace{0.1mm}
\includegraphics[width=.246\linewidth]{figs/openimages-cluster/32.jpg}
\vspace{0.1mm}
\includegraphics[width=.246\linewidth]{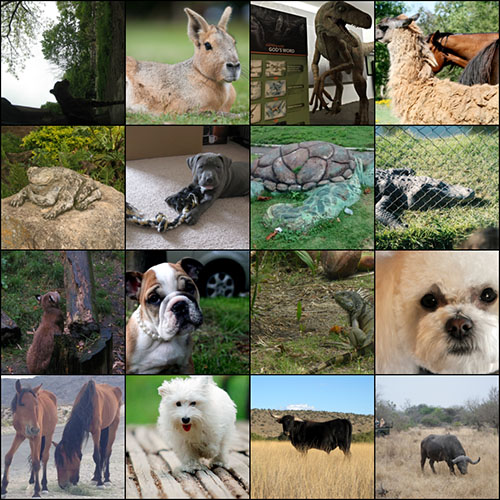}
\vspace{0.1mm}
\includegraphics[width=.246\linewidth]{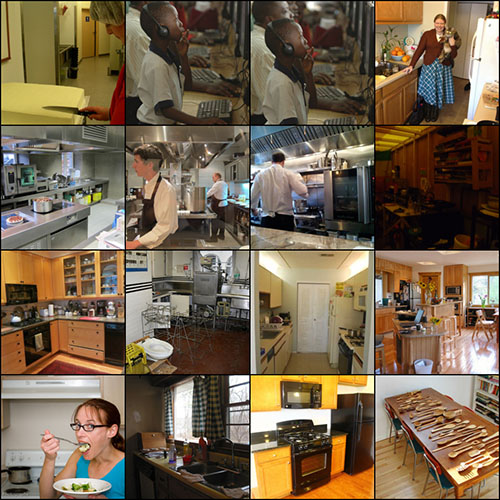}
\vspace{0.1mm}
\includegraphics[width=.246\linewidth]{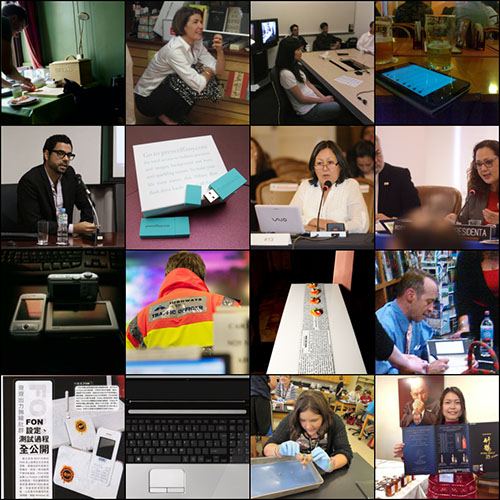}
\includegraphics[width=.246\linewidth]{figs/openimages-cluster/41.jpg}
\includegraphics[width=.246\linewidth]{figs/openimages-cluster/43.jpg}
\includegraphics[width=.246\linewidth]{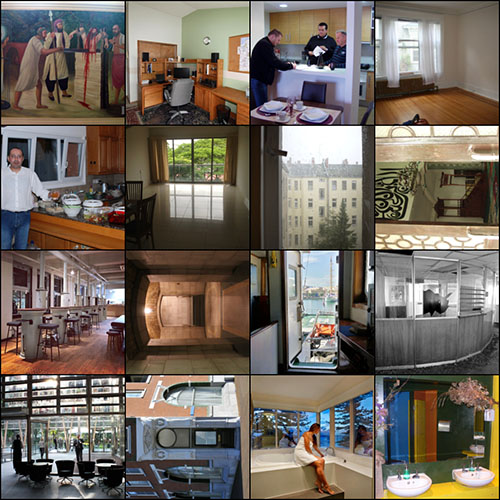}
\end{center}
\vspace{-5mm}
\caption{Top 8 clusters from COCO+OpenImages corresponding to the best performing expert adapted on PASCAL VOC.}
\label{fig:voc-clusters}
\end{figure*}

\begin{table}[t]
\resizebox{\linewidth}{!}{%
\addtolength{\tabcolsep}{-4pt}
\begin{tabular}{|ll|c|c|c|c|c|}
\hline
\multicolumn{2}{|c|}{\multirow{2}{*}{\textbf{Pretrain. Sel. Method}}} & \multicolumn{5}{c|}{\textbf{Target Dataset}} \\
\multicolumn{2}{|c|}{} & Stanf. Dogs & Stanf. Cars & Oxford-IIIT Pets & Flowers 102 & CUB200 Birds \\
\hline
0\% & Random Init. & 23.66 & 18.60 & 32.35 & 48.02 & 25.06 \\
\hline
100\% & Entire Dataset & 41.64 & 46.83 & 56.34 & 67.17 & 35.28 \\
\hline
\multirow{2}{*}{20\%} & Uniform Sample & 41.01 & 44.16 & 56.01 & 64.42 & 34.41 \\
 & NDS & 39.72 & 43.56 & 54.62 & 65.90 & 34.57 \\
\hline
\end{tabular}%
}
\vspace{-3mm}
\caption{MoCo Pretraining: Top-1 classification accuracy on five client datasets (columns) pretrained on the different subsets of data (rows) on the pretext task of instance discrimination.}
\label{table:moco}
\vspace{-5mm}
\end{table}

\begin{table}[t]
\resizebox{\linewidth}{!}{%
\addtolength{\tabcolsep}{-4pt}
\begin{tabular}{|ll|c|c|c|c|c|}
\hline
\multicolumn{2}{|c|}{\multirow{2}{*}{\textbf{Pretrain. Sel. Method}}} & \multicolumn{5}{c|}{\textbf{Target Dataset}} \\
\multicolumn{2}{|c|}{} & Stanf. Dogs & Stanf. Cars & Oxford-IIIT Pets & Flowers 102 & CUB200 Birds \\
\hline
0\% & Random Init. & 23.66 & 18.60 & 32.35 & 48.02 & 25.06 \\
\hline
100\% & Entire Dataset & 47.83 & 55.87 & 67.54 & 78.99 & 44.25 \\
\hline
\multirow{2}{*}{20\%} & Uniform Sample & 42.74 & 42.82 & 60.25 & 72.59 & 39.30 \\
 & NDS & 43.33 & 43.34 & 61.49 & 72.85 & 40.47 \\
\hline
\end{tabular}%
}
\vspace{-3mm}
\caption{RotNet Pretraining: Top-1 classification accuracy on five client datasets (columns) pretrained on the different subsets of data (rows) on the pretext task of predicting image rotations. }
\label{table:rotnet}
\vspace{-5mm}
\end{table}

\section{Appendix}
\vspace{-1mm}
In the Appendix, we provide additional details and results for our Neural Data Server. 
\vspace{-0mm}
\subsection{Web Interface}

Our NDS is running as a web-service at \url{http://aidemos.cs.toronto.edu/nds/}. We are inviting interested readers to try it and give us feedback. A snapshot of the website is provided in Figure~\ref{fig:web-interface}.

\subsection{Additional Results}
\vspace{0mm}
We visually assess domain confusion in Figures~\ref{fig:modanet-clusters},~\ref{fig:cityscapes-clusters}, and~\ref{fig:voc-clusters}. We randomly select 9 images per cluster and display the top 8 clusters corresponding to the experts with the highest proxy task performance in miniModaNet, Cityscapes, and VOC-Pascal. We can observe  that the images from the top clusters do indeed reflect the types of objects one encounters for autonomous driving, fashion, and general scenes corresponding to the respective target (client) datasets, showcasing the plausibility of our NDS. 

We further extend Table 2 and 3 in the main paper by showing detailed instance segmentation results for fine-tuning on the Cityscapes dataset. We report the performance measured by the COCO-style mask AP (averaged over IoU thresholds) for the 8 object categories. Table~\ref{table:maskrcnn-seg-coco} reports the mask AP by sampling 23K, 47K, and 59K images from COCO to be used for pre-training for Cityscapes, and Table~\ref{table:maskrcnn-seg-openimages} reports the mask AP by sampling 118K, 200K images from OpenImages for pre-training. 

\noindent\textbf{Self-Supervised Pretraining:} We evaluate NDS in a scenario where a client uses self-supervised learning to pretrain on the selected server data. We follow the same setup as described in Section 4.2, except that rather than pretraining using classification labels, clients ignore the availability of the labels and pretrain using two self-supervised learning approaches: MoCo~\cite{he2019momentum} and RotNet~\cite{gidaris2018unsupervised}. In Table ~\ref{table:moco}, we pretrain on the selected data subset using MoCo, an approach recently proposed by He \etal, where the model is trained on the pretext task of instance discrimination. In Table~\ref{table:rotnet}, we use ~\cite{gidaris2018unsupervised} to pretrain our model on the pretext task of predicting image rotation. We observe that in the case of MoCo, pretraining on NDS selected subset does not always yield better performance than pretraining on a randomly sampled subset of the same size. In the case of RotNet, pretraining on NDS selected subset has a slight gain over the baseline of uniform sampling. These results suggest that the optimal dataset for pretraining using self-supervised learning may be dependent on the pretext task. More formal studies on the relationship connecting training data, pretraining task, and transferring performance is required. 

{\small
\bibliographystyle{ieee_fullname}
\bibliography{egbib}

\begin{thebibliography}{10}\itemsep=-1pt

\bibitem{AbuElHaija2016YouTube8MAL}
Sami Abu-El-Haija, Nisarg Kothari, Joonseok Lee, Apostol Natsev, George
  Toderici, Balakrishnan Varadarajan, and Sudheendra Vijayanarasimhan.
\newblock Youtube-8m: A large-scale video classification benchmark.
\newblock {\em ArXiv}, abs/1609.08675, 2016.

\bibitem{Achille2019Task2VecTE}
Alessandro Achille, Michael Lam, Rahul Tewari, Avinash Ravichandran, Subhransu
  Maji, Charless~C. Fowlkes, Stefano Soatto, and Pietro Perona.
\newblock Task2vec: Task embedding for meta-learning.
\newblock {\em 2019 IEEE/CVF International Conference on Computer Vision
  (ICCV)}, pages 6429--6438, 2019.

\bibitem{Acuna_2019_CVPR}
David Acuna, Amlan Kar, and Sanja Fidler.
\newblock Devil is in the edges: Learning semantic boundaries from noisy
  annotations.
\newblock In {\em CVPR}, 2019.

\bibitem{acuna2018efficient}
David Acuna, Huan Ling, Amlan Kar, and Sanja Fidler.
\newblock Efficient interactive annotation of segmentation datasets with
  polygon-rnn++.
\newblock In {\em CVPR}, 2018.

\bibitem{BenDavid2009ATO}
Shai Ben-David, John Blitzer, Koby Crammer, Alex Kulesza, Fernando Pereira, and
  Jennifer~Wortman Vaughan.
\newblock A theory of learning from different domains.
\newblock {\em Machine Learning}, 79:151--175, 2009.

\bibitem{NIPS2006_2983}
Shai Ben-David, John Blitzer, Koby Crammer, and Fernando Pereira.
\newblock Analysis of representations for domain adaptation.
\newblock In B. Sch\"{o}lkopf, J.~C. Platt, and T. Hoffman, editors, {\em
  NeurIPS}, pages 137--144. MIT Press, 2007.

\bibitem{bengio2009curriculum}
Yoshua Bengio, J{\'e}r{\^o}me Louradour, Ronan Collobert, and Jason Weston.
\newblock Curriculum learning.
\newblock In {\em Proceedings of the 26th annual international conference on
  machine learning}, pages 41--48. ACM, 2009.

\bibitem{Bonawitz2017PracticalSA}
Keith Bonawitz, Vladimir Ivanov, Ben Kreuter, Antonio Marcedone, H.~Brendan
  McMahan, Sarvar Patel, Daniel Ramage, Aaron Segal, and Karn Seth.
\newblock Practical secure aggregation for privacy-preserving machine learning.
\newblock In {\em ACM Conf. on Computer and Communications Security}, 2017.

\bibitem{Caelles2018The2D}
Sergi Caelles, Alberto Montes, Kevis-Kokitsi Maninis, Yuhua Chen, Luc~Van Gool,
  Federico Perazzi, and Jordi Pont-Tuset.
\newblock The 2018 davis challenge on video object segmentation.
\newblock {\em ArXiv}, abs/1803.00557, 2018.

\bibitem{Caesar2019nuScenesAM}
Holger Caesar, Varun Bankiti, Alex~H. Lang, Sourabh Vora, Venice~Erin Liong,
  Qiang Xu, Anush Krishnan, Yu Pan, Giancarlo Baldan, and Oscar Beijbom.
\newblock nuscenes: A multimodal dataset for autonomous driving.
\newblock {\em ArXiv}, abs/1903.11027, 2019.

\bibitem{deeplab}
Liang-Chieh Chen, George Papandreou, Iasonas Kokkinos, Kevin Murphy, and
  Alan~L. Yuille.
\newblock Deeplab: Semantic image segmentation with deep convolutional nets,
  atrous convolution, and fully connected crfs.
\newblock {\em CoRR}, abs/1606.00915, 2016.

\bibitem{chen2015marginalizing}
Minmin Chen, Kilian~Q Weinberger, Zhixiang Xu, and Fei Sha.
\newblock Marginalizing stacked linear denoising autoencoders.
\newblock {\em Journal of Machine Learning Research}, 16(1):3849--3875, 2015.

\bibitem{chrabaszcz2017downsampled}
Patryk Chrabaszcz, Ilya Loshchilov, and Frank Hutter.
\newblock A downsampled variant of imagenet as an alternative to the cifar
  datasets, 2017.

\bibitem{Cordts2016TheCD}
Marius Cordts, Mohamed Omran, Sebastian Ramos, Timo Rehfeld, Markus Enzweiler,
  Rodrigo Benenson, Uwe Franke, Stefan Roth, and Bernt Schiele.
\newblock The cityscapes dataset for semantic urban scene understanding.
\newblock {\em CVPR}, pages 3213--3223, 2016.

\bibitem{csurka2017domain}
Gabriela Csurka.
\newblock Domain adaptation for visual applications: A comprehensive survey.
\newblock {\em arXiv preprint arXiv:1702.05374}, 2017.

\bibitem{Cui2018LargeSF}
Yin Cui, Yang Song, Chen Sun, Andrew Howard, and Serge~J. Belongie.
\newblock Large scale fine-grained categorization and domain-specific transfer
  learning.
\newblock {\em CVPR}, pages 4109--4118, 2018.

\bibitem{imagenet_cvpr09}
J. Deng, W. Dong, R. Socher, L.-J. Li, K. Li, and L. Fei-Fei.
\newblock {ImageNet: A Large-Scale Hierarchical Image Database}.
\newblock In {\em CVPR}, 2009.

\bibitem{Everingham2009ThePV}
Mark Everingham, Luc~Van Gool, Christopher K.~I. Williams, John~M. Winn, and
  Andrew Zisserman.
\newblock The pascal visual object classes (voc) challenge.
\newblock {\em International Journal of Computer Vision}, 88:303--338, 2009.

\bibitem{Ganin2015DomainAdversarialTO}
Yaroslav Ganin, Evgeniya Ustinova, Hana Ajakan, Pascal Germain, Hugo
  Larochelle, François Laviolette, Mario Marchand, and Victor~S. Lempitsky.
\newblock Domain-adversarial training of neural networks.
\newblock {\em J. Mach. Learn. Res.}, 17:59:1--59:35, 2015.

\bibitem{Geiger2012CVPR}
Andreas Geiger, Philip Lenz, and Raquel Urtasun.
\newblock Are we ready for autonomous driving? the kitti vision benchmark
  suite.
\newblock In {\em CVPR}, 2012.

\bibitem{gidaris2018unsupervised}
Spyros Gidaris, Praveer Singh, and Nikos Komodakis.
\newblock Unsupervised representation learning by predicting image rotations.
\newblock In {\em ICLR}, 2018.

\bibitem{gross2017hard}
Sam Gross, Marc'Aurelio Ranzato, and Arthur Szlam.
\newblock Hard mixtures of experts for large scale weakly supervised vision.
\newblock In {\em CVPR}, pages 6865--6873, 2017.

\bibitem{he2019momentum}
Kaiming He, Haoqi Fan, Yuxin Wu, Saining Xie, and Ross Girshick.
\newblock Momentum contrast for unsupervised visual representation learning,
  2019.

\bibitem{he2018rethinking}
Kaiming He, Ross~B. Girshick, and Piotr Doll{\'{a}}r.
\newblock Rethinking imagenet pre-training.
\newblock {\em CoRR}, abs/1811.08883, 2018.

\bibitem{He2017MaskR}
Kaiming He, Georgia Gkioxari, Piotr Doll{\'a}r, and Ross~B. Girshick.
\newblock Mask r-cnn.
\newblock {\em ICCV}, pages 2980--2988, 2017.

\bibitem{He2015DeepRL}
Kaiming He, Xiangyu Zhang, Shaoqing Ren, and Jian Sun.
\newblock Deep residual learning for image recognition.
\newblock {\em CVPR}, pages 770--778, 2015.

\bibitem{Hinton2015DistillingTK}
Geoffrey~E. Hinton, Oriol Vinyals, and Jeffrey Dean.
\newblock Distilling the knowledge in a neural network.
\newblock {\em ArXiv}, abs/1503.02531, 2015.

\bibitem{Jacobs1991AdaptiveMO}
Robert~A. Jacobs, Michael~I. Jordan, Steven~J. Nowlan, and Geoffrey~E. Hinton.
\newblock Adaptive mixtures of local experts.
\newblock {\em Neural Computation}, 3:79--87, 1991.

\bibitem{Metasim19}
Amlan Kar, Aayush Prakash, Ming-Yu Liu, Eric Cameracci, Justin Yuan, Matt
  Rusiniak, David Acuna, Antonio Torralba, and Sanja Fidler.
\newblock Meta-sim: Learning to generate synthetic datasets.
\newblock In {\em ICCV}, 2019.

\bibitem{KhoslaYaoJayadevaprakashFeiFei_FGVC2011}
Aditya Khosla, Nityananda Jayadevaprakash, Bangpeng Yao, and Li Fei-Fei.
\newblock Novel dataset for fine-grained image categorization.
\newblock In {\em CVPR Workshop on Fine-Grained Visual Categorization},
  Colorado Springs, CO, June 2011.

\bibitem{KrauseStarkDengFei-Fei_3DRR2013}
Jonathan Krause, Michael Stark, Jia Deng, and Li Fei-Fei.
\newblock 3d object representations for fine-grained categorization.
\newblock In {\em IEEE Workshop on 3D Representation and Recognition
  (3dRR-13)}, Sydney, Australia, 2013.

\bibitem{OpenImages}
Alina Kuznetsova, Hassan Rom, Neil Alldrin, Jasper Uijlings, Ivan Krasin, Jordi
  Pont-Tuset, Shahab Kamali, Stefan Popov, Matteo Malloci, Tom Duerig, and
  Vittorio Ferrari.
\newblock The open images dataset v4: Unified image classification, object
  detection, and visual relationship detection at scale.
\newblock {\em arXiv:1811.00982}, 2018.

\bibitem{Li2019AnAO}
Hengduo Li, Bharat Singh, Mahyar Najibi, Zuxuan Wu, and Larry~S. Davis.
\newblock An analysis of pre-training on object detection.
\newblock {\em ArXiv}, abs/1904.05871, 2019.

\bibitem{Lin2014MicrosoftCC}
Tsung-Yi Lin, Michael Maire, Serge~J. Belongie, Lubomir~D. Bourdev, Ross~B.
  Girshick, James Hays, Pietro Perona, Deva Ramanan, Piotr Doll{\'a}r, and
  C.~Lawrence Zitnick.
\newblock Microsoft coco: Common objects in context.
\newblock In {\em ECCV}, 2014.

\bibitem{mahajan2018exploring}
Dhruv Mahajan, Ross Girshick, Vignesh Ramanathan, Kaiming He, Manohar Paluri,
  Yixuan Li, Ashwin Bharambe, and Laurens van~der Maaten.
\newblock Exploring the limits of weakly supervised pretraining.
\newblock In {\em ECCV}, pages 181--196, 2018.

\bibitem{McMahan2016FederatedLO}
H.~Brendan McMahan, Eider Moore, Daniel Ramage, and Blaise~Ag{\"u}era y Arcas.
\newblock Federated learning of deep networks using model averaging.
\newblock {\em ArXiv}, abs/1602.05629, 2016.

\bibitem{mehta2019active}
Bhairav Mehta, Manfred Diaz, Florian Golemo, Christopher~J Pal, and Liam Paull.
\newblock Active domain randomization.
\newblock {\em arXiv preprint arXiv:1904.04762}, 2019.

\bibitem{ngiam2018domain}
Jiquan Ngiam, Daiyi Peng, Vijay Vasudevan, Simon Kornblith, Quoc~V. Le, and
  Ruoming Pang.
\newblock Domain adaptive transfer learning with specialist models, 2018.

\bibitem{Nilsback08}
M-E. Nilsback and A. Zisserman.
\newblock Automated flower classification over a large number of classes.
\newblock In {\em Proc. of the Indian Conference on Computer Vision, Graphics
  and Image Processing}, Dec 2008.

\bibitem{pan2009survey}
Sinno~Jialin Pan and Qiang Yang.
\newblock A survey on transfer learning.
\newblock {\em IEEE Trans. on knowledge and data engineering},
  22(10):1345--1359, 2009.

\bibitem{parkhi12a}
O.~M. Parkhi, A. Vedaldi, A. Zisserman, and C.~V. Jawahar.
\newblock Cats and dogs.
\newblock In {\em CVPR}, 2012.

\bibitem{ruiz2018learning}
Nataniel Ruiz, Samuel Schulter, and Manmohan Chandraker.
\newblock Learning to simulate.
\newblock {\em arXiv preprint arXiv:1810.02513}, 2018.

\bibitem{settles2009active}
Burr Settles.
\newblock Active learning literature survey.
\newblock Technical report, University of Wisconsin-Madison Department of
  Computer Sciences, 2009.

\bibitem{Shelhamer2017FCN}
Evan Shelhamer, Jonathan Long, and Trevor Darrell.
\newblock Fully convolutional networks for semantic segmentation.
\newblock {\em PAMI}, 39(4):640--651, Apr. 2017.

\bibitem{sun2017revisiting}
Chen Sun, Abhinav Shrivastava, Saurabh Singh, and Abhinav Gupta.
\newblock Revisiting unreasonable effectiveness of data in deep learning era.
\newblock In {\em ICCV}, pages 843--852, 2017.

\bibitem{Takikawa2019GatedSCNNGS}
Towaki Takikawa, David Acuna, Varun Jampani, and Sanja Fidler.
\newblock Gated-scnn: Gated shape cnns for semantic segmentation.
\newblock {\em ArXiv}, abs/1907.05740, 2019.

\bibitem{tremblay2018training}
Jonathan Tremblay, Aayush Prakash, David Acuna, Mark Brophy, Varun Jampani, Cem
  Anil, Thang To, Eric Cameracci, Shaad Boochoon, and Stan Birchfield.
\newblock Training deep networks with synthetic data: Bridging the reality gap
  by domain randomization.
\newblock In {\em CVPR Workshop}, pages 969--977, 2018.

\bibitem{tripathi2019learning}
Shashank Tripathi, Siddhartha Chandra, Amit Agrawal, Ambrish Tyagi, James~M
  Rehg, and Visesh Chari.
\newblock Learning to generate synthetic data via compositing.
\newblock In {\em CVPR}, pages 461--470, 2019.

\bibitem{WahCUB_200_2011}
C. Wah, S. Branson, P. Welinder, P. Perona, and S. Belongie.
\newblock {The Caltech-UCSD Birds-200-2011 Dataset}.
\newblock Technical Report CNS-TR-2011-001, California Institute of Technology,
  2011.

\bibitem{Yosinski2014HowTA}
Jason Yosinski, Jeff Clune, Yoshua Bengio, and Hod Lipson.
\newblock How transferable are features in deep neural networks?
\newblock In {\em NeurIPS}, 2014.

\bibitem{Zamir2018TaskonomyDT}
Amir~Roshan Zamir, Alexander Sax, William~B. Shen, Leonidas~J. Guibas,
  Jagannath Malik, and Silvio Savarese.
\newblock Taskonomy: Disentangling task transfer learning.
\newblock {\em CVPR}, pages 3712--3722, 2018.

\bibitem{Zheng2018ModaNetAL}
Shuai Zheng, Fan Yang, M.~Hadi Kiapour, and Robinson Piramuthu.
\newblock Modanet: A large-scale street fashion dataset with polygon
  annotations.
\newblock In {\em ACM Multimedia}, 2018.

\bibitem{Zhu2018ImprovingSS}
Yi Zhu, Karan Sapra, Fitsum~A. Reda, Kevin~J. Shih, Shawn~D. Newsam, Andrew
  Tao, and Bryan Catanzaro.
\newblock Improving semantic segmentation via video propagation and label
  relaxation.
\newblock In {\em CVPR}, 2018.

\end{thebibliography}
}

\end{document}